\documentclass[10pt,twocolumn,letterpaper]{article}

\usepackage{cvpr,times,graphicx,tabulary,multirow,subfig,xspace,xcolor}
\usepackage{amsthm,amssymb,fixmath,mathtools,nicefrac}
\usepackage[pagebackref=true,breaklinks=true,letterpaper=true,colorlinks,
  citecolor=citecolor,bookmarks=false]{hyperref}
\definecolor{citecolor}{RGB}{34,139,34}
\def\cvprPaperID{}\cvprfinalcopy

\newcommand{\app}{\raise.17ex\hbox{$\scriptstyle\sim$}}
\newcommand{\tss}[1]{\textsuperscript{#1}}

\newcommand{\dt}[1]{\fontsize{8pt}{.1em}\selectfont \emph{#1}}
\newcommand{\bd}[1]{\textbf{#1}}
\newcommand{\x}{$\times$}

\newlength\savewidth\newcommand\shline{\noalign{\global\savewidth\arrayrulewidth
  \global\arrayrulewidth 1pt}\hline\noalign{\global\arrayrulewidth\savewidth}}
\newcolumntype{x}[1]{>{\centering\arraybackslash}p{#1pt}}
\newcommand{\tablestyle}[2]{\setlength{\tabcolsep}{#1}\renewcommand{\arraystretch}{#2}\centering\small}

\makeatletter
\renewcommand\paragraph{\@startsection{paragraph}{4}{\z@}%
  {.5em \@plus1ex \@minus.1ex}%
  {-.5em}%
  {\normalfont\normalsize\bfseries}}
\makeatother

\newcommand{\things}{\tss{Th}\xspace}
\newcommand{\stuff}{\tss{St}\xspace}

\begin{document}
\title{Panoptic Feature Pyramid Networks\vspace{-4mm}}
\author{%
 Alexander Kirillov \quad Ross Girshick \quad Kaiming He \quad Piotr Doll\'ar\\[2mm]
 Facebook AI Research (FAIR)}
\maketitle

%%%%%%%%%%%%%%%%%%%%%%%%%%%%%%%%%%%%%%%%%%%%%%%%%%%%%%%%%%%%%%%%%%%%%%%%%%%%%%%%%%%%%%%%%%%%%%%%%%%
\begin{abstract}
The recently introduced panoptic segmentation task has renewed our community's interest in unifying the tasks of instance segmentation (for thing classes) and semantic segmentation (for stuff classes). However, current state-of-the-art methods for this joint task use separate and dissimilar networks for instance and semantic segmentation, without performing any shared computation. In this work, we aim to unify these methods at the \emph{architectural level}, designing a \emph{single network} for both tasks. Our approach is to endow Mask R-CNN, a popular instance segmentation method, with a semantic segmentation branch using a shared Feature Pyramid Network (FPN) backbone. Surprisingly, this \emph{simple baseline} not only remains effective for instance segmentation, but also yields a lightweight, top-performing method for semantic segmentation. In this work, we perform a detailed study of this \emph{minimally extended} version of Mask R-CNN with FPN, which we refer to as \emph{Panoptic FPN}, and show it is a robust and accurate baseline for both tasks. Given its effectiveness and conceptual simplicity, we hope our method can serve as a strong baseline and aid future research in panoptic segmentation.
\end{abstract}

%%%%%%%%%%%%%%%%%%%%%%%%%%%%%%%%%%%%%%%%%%%%%%%%%%%%%%%%%%%%%%%%%%%%%%%%%%%%%%%%%%%%%%%%%%%%%%%%%%%
\section{Introduction}

Our community has witnessed rapid progress in \emph{semantic segmentation}, where the task is to assign each pixel a class label (\eg for stuff classes), and more recently in \emph{instance segmentation}, where the task is to detect and segment each object instance (\eg for thing classes). These advances have been aided by simple yet powerful baseline methods, including Fully Convolutional Networks (FCN)~\cite{long2015fully} and Mask R-CNN~\cite{he2017mask} for semantic and instance segmentation, respectively. These methods are conceptually simple, fast, and flexible, serving as a foundation for much of the subsequent progress in these areas. In this work our goal is to propose a similarly simple, single-network baseline for the joint task of \emph{panoptic segmentation}~\cite{kirillov2017panoptic}, a task which encompasses both semantic and instance segmentation.

%##############################################################################################
\begin{figure}\centering
\includegraphics[width=.9\linewidth]{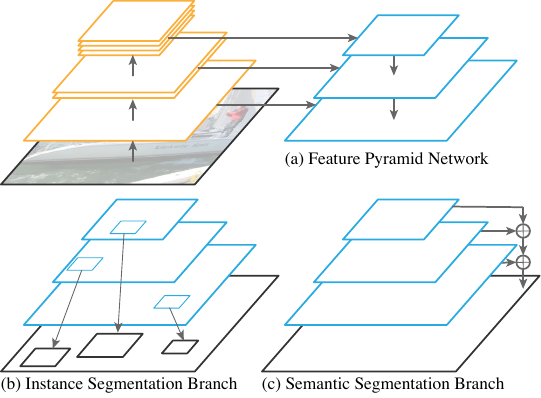}
\caption{\textbf{Panoptic FPN}: (a) We start with an FPN backbone~\cite{lin2016feature}, widely used in object detection, for extracting rich multi-scale features. (b) As in Mask R-CNN~\cite{he2017mask}, we use a region-based branch on top of FPN for instance segmentation. (c) In parallel, we add a lightweight dense-prediction branch on top of the same FPN features for semantic segmentation. This simple extension of Mask R-CNN with FPN is a fast and accurate baseline for both tasks.}\vspace{-2mm}
\label{fig:teaser}
\end{figure}
%##############################################################################################

While conceptually straightforward, designing a single network that achieves high accuracy for both tasks is challenging as top-performing methods for the two tasks have many differences. For semantic segmentation, FCNs with specialized backbones enhanced by dilated convolutions~\cite{yu2015multi,deeplabV2} dominate popular leaderboards \cite{everingham2015pascal, Cordts2016Cityscapes}. For instance segmentation, the region-based Mask R-CNN~\cite{he2017mask} with a Feature Pyramid Network (FPN)~\cite{lin2016feature} backbone has been used as a foundation for all top entries in recent recognition challenges \cite{lin2014coco, zhou2017ade20k, neuhold2017mapillary}. While there have been attempts to unify semantic and instance segmentation \cite{pham2017biseg, arnab2017pixelwise, chen2017masklab}, the specialization currently necessary to achieve top performance in each was perhaps inevitable given their parallel development and separate benchmarks.

%##############################################################################################
\begin{figure*}
\includegraphics[height=3.082cm]{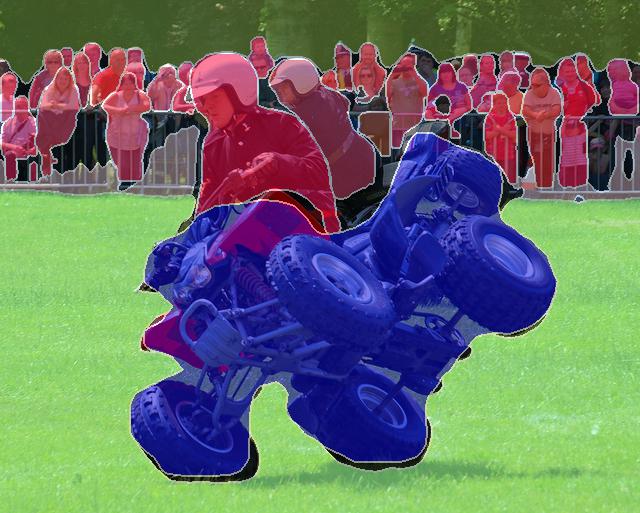}
\includegraphics[height=3.082cm]{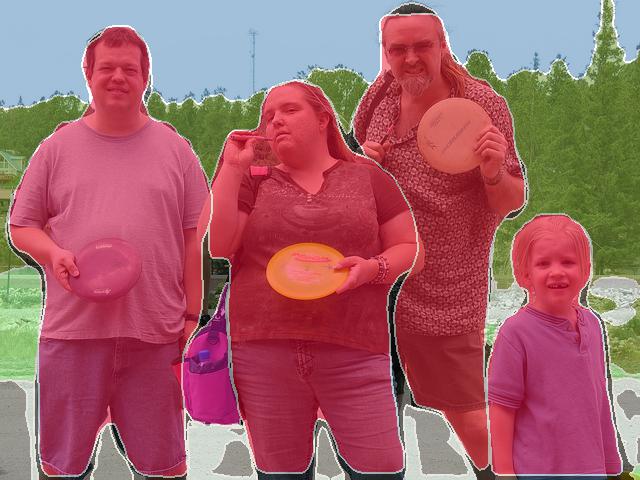}
\includegraphics[height=3.082cm]{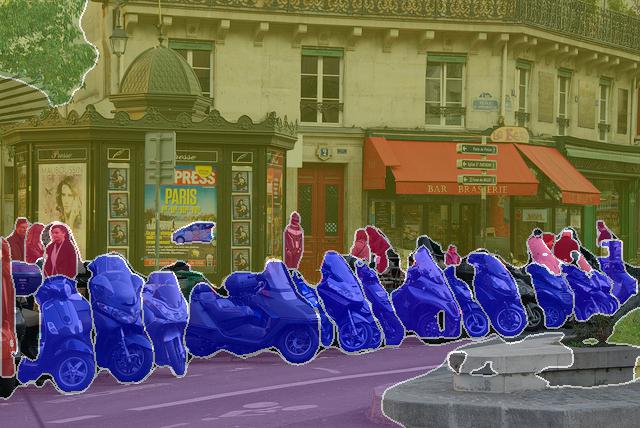}
\includegraphics[height=3.082cm]{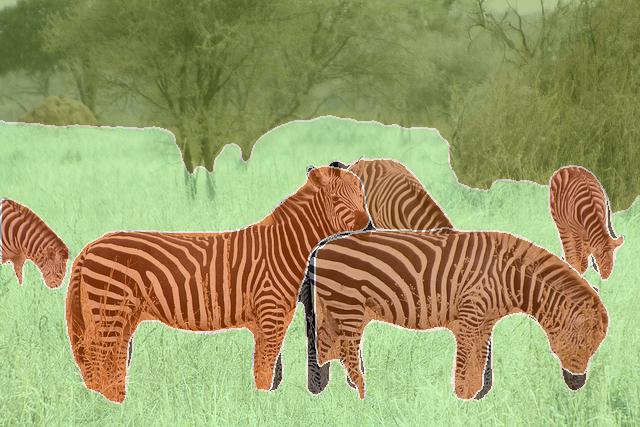}\\[.5mm]
\includegraphics[width=0.33\linewidth]{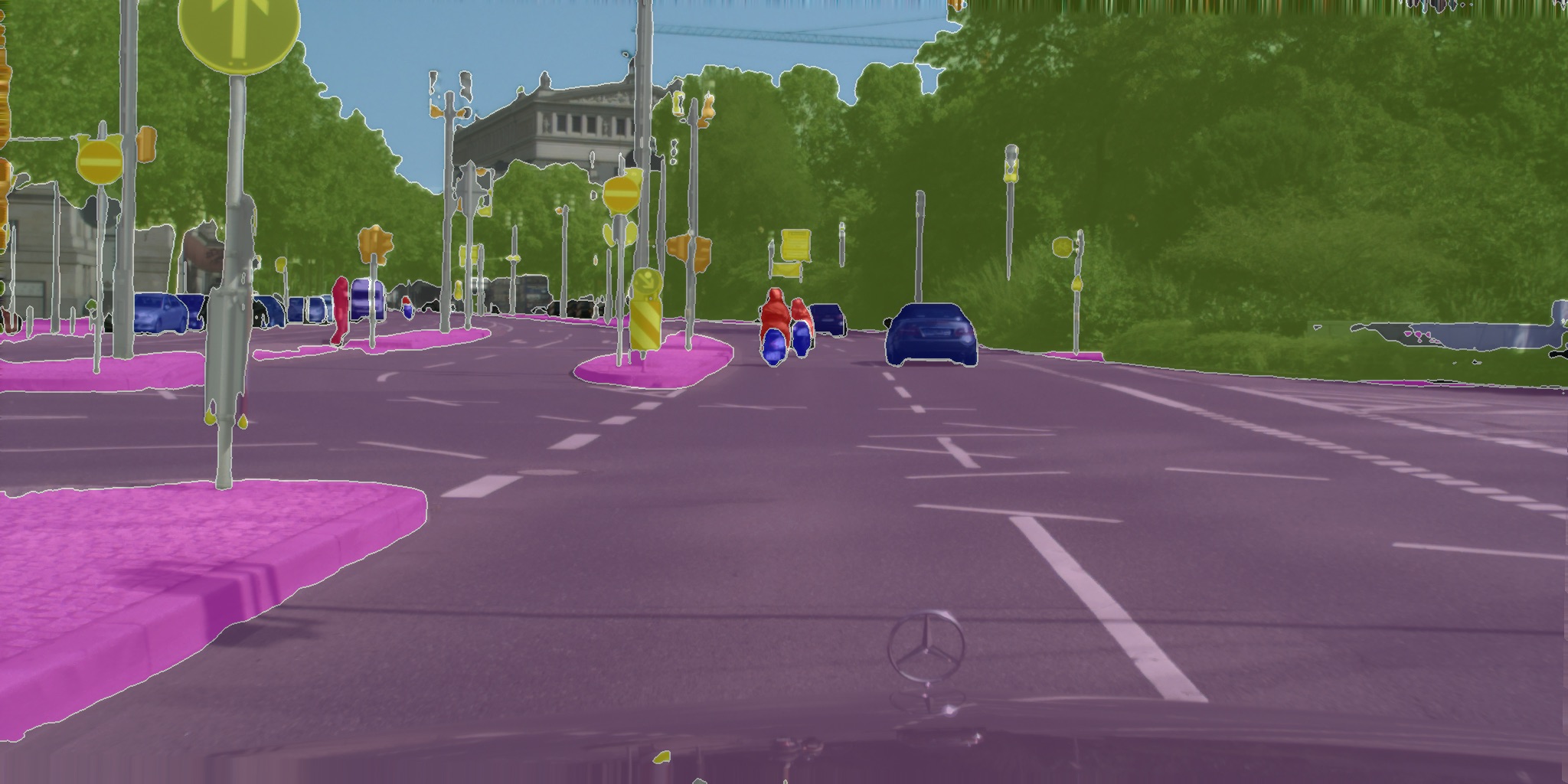}
\includegraphics[width=0.33\linewidth]{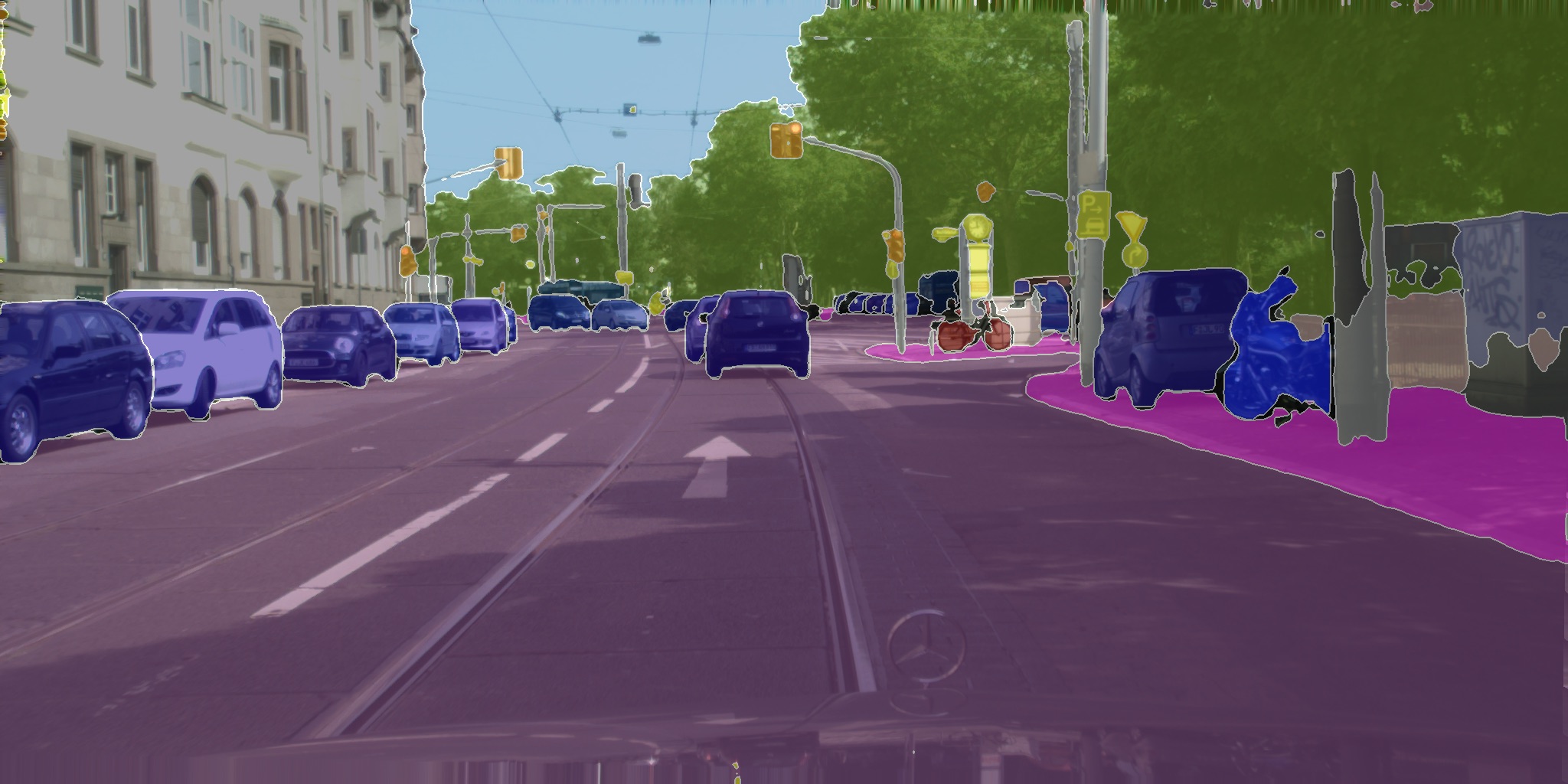}
\includegraphics[width=0.33\linewidth]{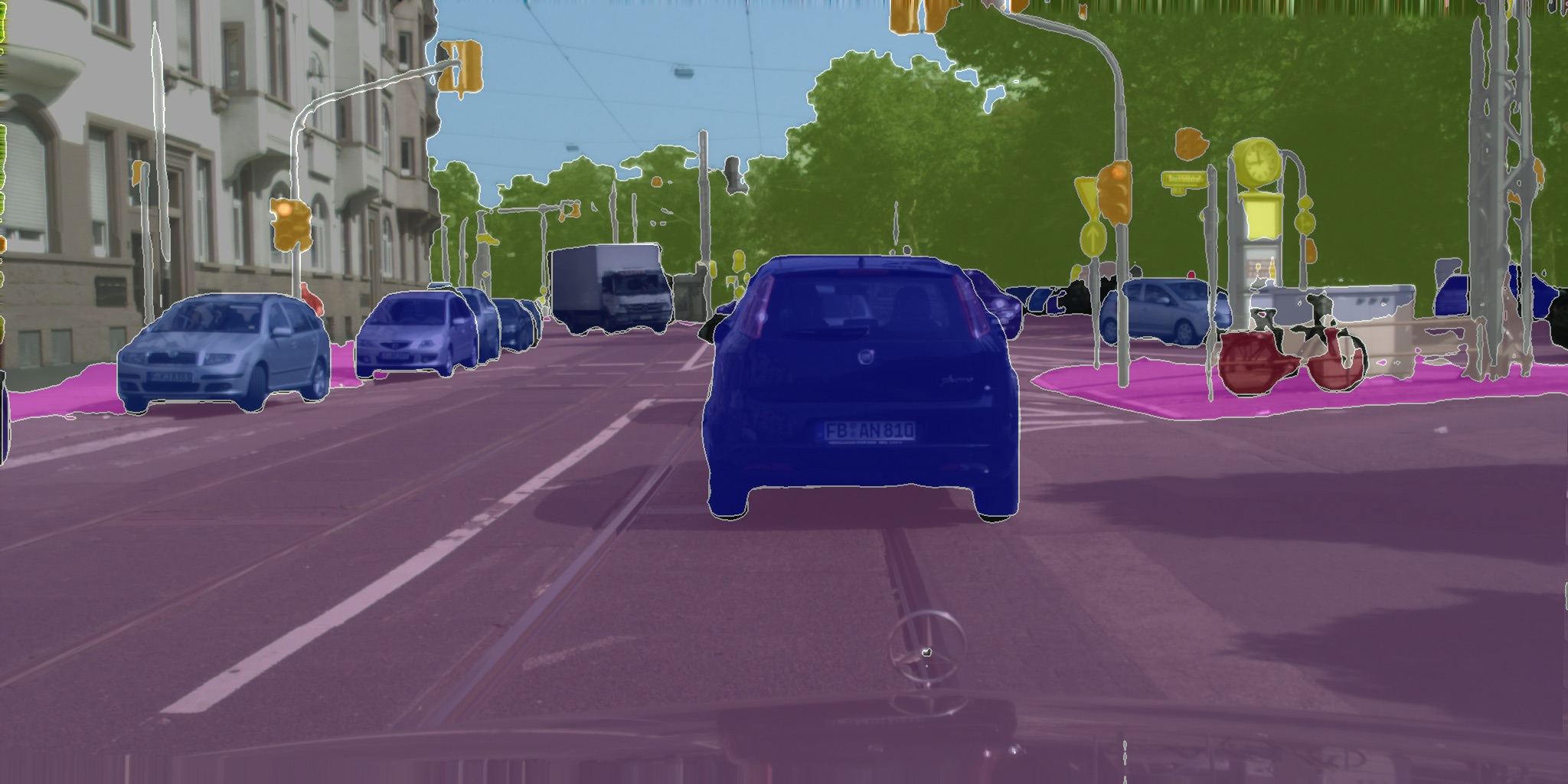}
\caption{Panoptic FPN results on COCO (top) and Cityscapes (bottom) using a single ResNet-101-FPN network.}
\label{fig:results:1}\vspace{-3mm}
\end{figure*}
%##############################################################################################

Given the architectural differences in these top methods, one might expect compromising accuracy on either instance or semantic segmentation is necessary when designing a single network for both tasks. Instead, we show a simple, flexible, and effective architecture that can match accuracy for both tasks using \emph{a single network that simultaneously generates region-based outputs (for instance segmentation) and dense-pixel outputs (for semantic segmentation)}.

Our approach starts with the FPN~\cite{lin2016feature} backbone popular for instance-level recognition~\cite{he2017mask} and adds a branch for performing semantic segmentation in parallel with the existing region-based branch for instance segmentation, see Figure~\ref{fig:teaser}. We make no changes to the FPN backbone when adding the dense-prediction branch, making it compatible with existing instance segmentation methods. Our method, which we call \emph{Panoptic FPN} for its ability to generate both instance and semantic segmentations via FPN, is easy to implement given the Mask R-CNN framework~\cite{Detectron2018}.

While Panoptic FPN is an intuitive extension of Mask R-CNN with FPN, properly training the two branches for simultaneous region-based and dense-pixel prediction is important for good results. We perform careful studies in the joint setting for how to balance the losses for the two branches, construct minibatches effectively, adjust learning rate schedules, and perform data augmentation. We also explore various designs for the semantic segmentation branch (all other network components follow Mask R-CNN). Overall, while our approach is robust to exact design choices, properly addressing these issues is key for good results.

When trained for each task independently, our method achieves excellent results for both instance and semantic segmentation on both COCO~\cite{lin2014coco} and Cityscapes~\cite{Cordts2016Cityscapes}. For instance segmentation, this is expected as our method in this case is equivalent to Mask R-CNN\@. For semantic segmentation, our simple dense-prediction branch attached to FPN yields accuracy on par with the latest dilation-based methods, such as the recent DeepLabV3+~\cite{deeplabV3plus}.

For panoptic segmentation~\cite{kirillov2017panoptic}, we demonstrate that with proper training, \emph{using a single FPN for solving both tasks simultaneously yields accuracy equivalent to training two separate FPNs}, with roughly half the compute. With the \emph{same} compute, a joint network for the two tasks outperforms two independent networks by a healthy margin. Example panoptic segmentation results are shown in Fig.~\ref{fig:results:1}.

Panoptic FPN is memory and computationally efficient, incurring only a slight overhead over Mask R-CNN\@. By avoiding the use of dilation, which has high overhead, our method can use any standard top-performing backbone (\eg a large ResNeXt~\cite{xie2017aggregated}). We believe this flexibility, together with the fast training and inference speeds of our method, will benefit future research on panoptic segmentation.

We used a preliminary version of our model (semantic segmentation branch only) as the foundation of the first-place winning entry in the COCO Stuff Segmentation~\cite{caesar2016coco} track in 2017. This single-branch model has since been adopted and generalized by several entries in the 2018 COCO and Mapillary Challenges$^{\ref{footnote:workshop}}$, showing its flexibility and effectiveness. We hope our proposed joint panoptic segmentation baseline is similarly impactful.

%%%%%%%%%%%%%%%%%%%%%%%%%%%%%%%%%%%%%%%%%%%%%%%%%%%%%%%%%%%%%%%%%%%%%%%%%%%%%%%%%%%%%%%%%%%%%%%%%%%
\section{Related Work}

\paragraph{Panoptic segmentation:} The joint task of thing and stuff segmentation has a rich history, including early work on scene parsing~\cite{tighe2014scene}, image parsing~\cite{tu2005image}, and holistic scene understanding~\cite{yao2012describing}. With the recent introduction of the joint \emph{panoptic segmentation} task~\cite{kirillov2017panoptic}, which includes a simple task specification and carefully designed task metrics, there has been a renewed interest in the joint task.

This year's COCO and Mapillary Recognition Challenge~\cite{lin2014coco, neuhold2017mapillary} featured panoptic segmentation tracks that proved popular. However, every competitive entry in the panoptic challenges used \emph{separate networks for instance and semantic segmentation}, with no shared computation.\footnote{\label{footnote:workshop}For details of not yet published winning entries in the 2018 COCO and Mapillary Recognition Challenge please see: \url{http://cocodataset.org/workshop/coco-mapillary-eccv-2018.html}. {\tt TRI-ML} used separate networks for the challenge but a joint network in their recent updated tech report~\cite{li2018learning} (which cites a preliminary version of our work).} Our goal is to design a \emph{single network} effective for both tasks that can serve as a baseline for future work.

\paragraph{Instance segmentation:} Region-based approaches to object detection, including the Slow/Fast/Faster/Mask R-CNN family~\cite{girshick2014rcnn, girshick2015fast, ren2015a, he2017mask}, which apply deep networks on candidate object regions, have proven highly successful. All recent winners of the COCO detection challenges have built on Mask R-CNN~\cite{he2017mask} with FPN~\cite{lin2016feature}, including in 2017~\cite{liu2018path, peng2018megdet} and 2018.$^{\ref{footnote:workshop}}$ Recent innovations include Cascade R-CNN~\cite{cai2018cascade}, deformable convolution~\cite{dai2017deformable}, and sync batch norm~\cite{peng2018megdet}. In this work, the original Mask R-CNN with FPN serves as the starting point for our baseline, giving us excellent instance segmentation performance, and making our method fully compatible with these recent advances.

An alternative to region-based instance segmentation is to start with a pixel-wise semantic segmentation and then perform grouping to extract instances~\cite{kirillov2016instancecut, liu2017sgn, arnab2017pixelwise}. This direction is innovative and promising. However, these methods tend to use \emph{separate networks} to predict the instance-level information (\eg, \cite{kirillov2016instancecut, arnab2017pixelwise, liu2017sgn} use a separate network to predict instance edges, bounding boxes, and object breakpoints, respectively). Our goal is to design a \emph{single network} for the joint task. Another interesting direction is to use position-sensitive pixel labeling~\cite{li2016fully} to encode instance information fully convolutionally; \cite{pham2017biseg, chen2017masklab} build on this.

Nevertheless, region-based approaches remain dominant on detection leaderboards~\cite{lin2014coco, zhou2017ade20k, neuhold2017mapillary}. While this motivates us to start with a region-based approach to instance segmentation, our approach would be fully compatible with a dense-prediction branch for instance segmentation.

\paragraph{Semantic segmentation:} FCNs~\cite{long2015fully} serve as the foundation of modern semantic segmentation methods. To increase feature resolution, which is necessary for generating high-quality results, recent top methods~\cite{deeplabV3plus, zhao2017pspnet, bulo2017place, zhao2018psanet} rely heavily on the use of dilated convolution~\cite{yu2015multi} (also known as atrous convolution~\cite{deeplabV2}). While effective, such an approach can substantially increase compute and memory, limiting the type of backbone network that can be used. To keep this flexibility, and more importantly to maintain compatibility with Mask R-CNN, we opt for a different approach.

As an alternative to dilation, an encoder-decoder~\cite{badrinarayanan2015segnet} or `U-Net'~\cite{ronneberger2015u} architecture can be used to increase feature resolution~\cite{honari2016recombinator, newell2016stacked, ghiasi2016laplacian, pinheiro2016learning}. Encoder-decoders progressively upsample and combine high-level features from a feedforward network with features from lower-levels, ultimately generating semantically meaningful, high-resolution features (see Figure \ref{fig:backbone}). While dilated networks are currently more popular and dominate leaderboards, encoder-decoders have also been used for semantic segmentation~\cite{ronneberger2015u, badrinarayanan2015segnet, ghiasi2016laplacian}.

In our work we adopt an encoder-decoder framework, namely FPN~\cite{lin2016feature}. In contrast to `symmetric' decoders~\cite{ronneberger2015u}, FPN uses a lightweight decoder (see Fig.~\ref{fig:backbone}). FPN was designed for instance segmentation, and it serves as the default backbone for Mask R-CNN. We show that \emph{without changes, FPN can also be highly effective for semantic segmentation}.

\paragraph{Multi-task learning:} Our approach is related to multi-task learning. In general, using a single network to solve multiple diverse tasks degrades performance~\cite{kokkinos2016ubernet}, but various strategies can mitigate this~\cite{kendall2017multi,misra2016cross}. For related tasks, there can be gains from multi-task learning, \eg the box branch in Mask R-CNN benefits from the mask branch~\cite{he2017mask}, and joint detection and semantic segmentation of thing classes also shows gains~\cite{bell2016inside, cao2018triply, dvornik2017blitznet, pham2017biseg}. Our work studies the benefits of multi-task training for stuff and thing segmentation.

%%%%%%%%%%%%%%%%%%%%%%%%%%%%%%%%%%%%%%%%%%%%%%%%%%%%%%%%%%%%%%%%%%%%%%%%%%%%%%%%%%%%%%%%%%%%%%%%%%%
\section{Panoptic Feature Pyramid Network}

Our approach, Panoptic FPN, is a simple, \emph{single-network baseline} whose goal is to achieve top performance on both instance and semantic segmentation, and their joint task: panoptic segmentation~\cite{kirillov2017panoptic}. Our design principle is to start from Mask R-CNN with FPN, a strong instance segmentation baseline, and make \emph{minimal} changes to also generate a semantic segmentation dense-pixel output (see Figure~\ref{fig:teaser}).

\subsection{Model Architecture}

\paragraph{Feature Pyramid Network:} We begin by briefly reviewing FPN~\cite{lin2016feature}. FPN takes a standard network with features at multiple spatial resolutions (\eg, ResNet~\cite{he2016deep}), and adds a light top-down pathway with lateral connections, see Figure~\ref{fig:teaser}a. The top-down pathway starts from the deepest layer of the network and progressively upsamples it while adding in transformed versions of higher-resolution features from the bottom-up pathway. FPN generates a \emph{pyramid}, typically with scales from 1/32 to 1/4 resolution, where each pyramid level has the \emph{same channel dimension} (256 by default).

\paragraph{Instance segmentation branch:} The design of FPN, and in particular the use of the same channel dimension for all pyramid levels, makes it easy to attach a region-based object detector like Faster R-CNN~\cite{ren2015a}. Faster R-CNN performs region of interest (RoI) pooling on different pyramid levels and applies a shared network branch to predict a refined box and class label for each region. To output instance segmentations, we use Mask R-CNN~\cite{he2017mask}, which extends Faster R-CNN by adding an FCN branch to predict a binary segmentation mask for each candidate region, see Figure~\ref{fig:teaser}b.

\paragraph{Panoptic FPN:} As discussed, our approach is to modify Mask R-CNN with FPN to enable pixel-wise semantic segmentation prediction. However, to achieve accurate predictions, the features used for this task should: (1) be of suitably high resolution to capture fine structures, (2) encode sufficiently rich semantics to accurately predict class labels, and (3) capture multi-scale information to predict stuff regions at multiple resolutions. Although FPN was designed for object detection, these requirements -- \emph{high-resolution, rich, multi-scale features} -- identify exactly the characteristics of FPN. We thus propose to attach to FPN a simple and fast semantic segmentation branch, described next.

%##############################################################################################
\begin{figure}
\includegraphics[width=0.9\linewidth]{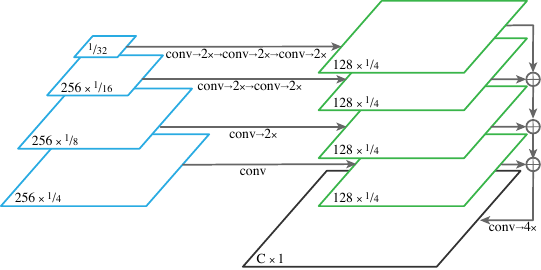}
\caption{\textbf{Semantic segmentation branch.} Each FPN level (left) is upsampled by convolutions and bilinear upsampling until it reaches 1/4 scale (right), theses outputs are then summed and finally transformed into a pixel-wise output.\vspace{-1mm}}\label{fig:semantic-branch}
\end{figure}
%##############################################################################################

\paragraph{Semantic segmentation branch:} To generate the semantic segmentation output from the FPN features, we propose a simple design to merge the information from all levels of the FPN pyramid into a single output. It is illustrated in detail in Figure~\ref{fig:semantic-branch}. Starting from the deepest FPN level (at 1/32 scale), we perform three upsampling stages to yield a feature map at 1/4 scale, where each upsampling stage consists of 3\x3 convolution, group norm~\cite{wu2018group}, ReLU, and 2\x~bilinear upsampling. This strategy is repeated for FPN scales 1/16, 1/8, and 1/4 (with progressively fewer upsampling stages). The result is a set of feature maps at the same 1/4 scale, which are then element-wise summed. A final 1\x1 convolution, 4\x~bilinear upsampling, and softmax are used to generate the per-pixel class labels at the original image resolution. In addition to stuff classes, this branch also outputs a special `other' class for all pixels belonging to objects (to avoid predicting stuff classes for such pixels).

\paragraph{Implementation details:} We use a standard FPN configuration with 256 output channels per scale, and our semantic segmentation branch reduces this to 128 channels. For the (pre-FPN) backbone, we use ResNet/ResNeXt~\cite{he2016deep,xie2017aggregated} models pre-trained on ImageNet~\cite{Russakovsky2015} using batch norm (BN)~\cite{ioffe2015batch}. When used in fine-tuning, we replace BN with a fixed channel-wise affine transformation, as is typical~\cite{he2016deep}.

\subsection{Inference and Training}\label{sec:inference-and-training}

\paragraph{Panoptic inference:} The panoptic output format~\cite{kirillov2017panoptic} requires each output pixel to be assigned a single class label (or void) and instance id (the instance id is ignored for stuff classes). As the instance and semantic segmentation outputs from Panoptic FPN may overlap; we apply the simple post-processing proposed in~\cite{kirillov2017panoptic} to resolve all overlaps. This post-processing is similar in spirit to non-maximum suppression and operates by: (1) resolving overlaps between different instances based on their confidence scores, (2) resolving overlaps between instance and semantic segmentation outputs in favor of instances, and (3) removing any stuff regions labeled `other' or under a given area threshold.

%##############################################################################################
\begin{figure}\centering
\includegraphics[width=.495\linewidth]{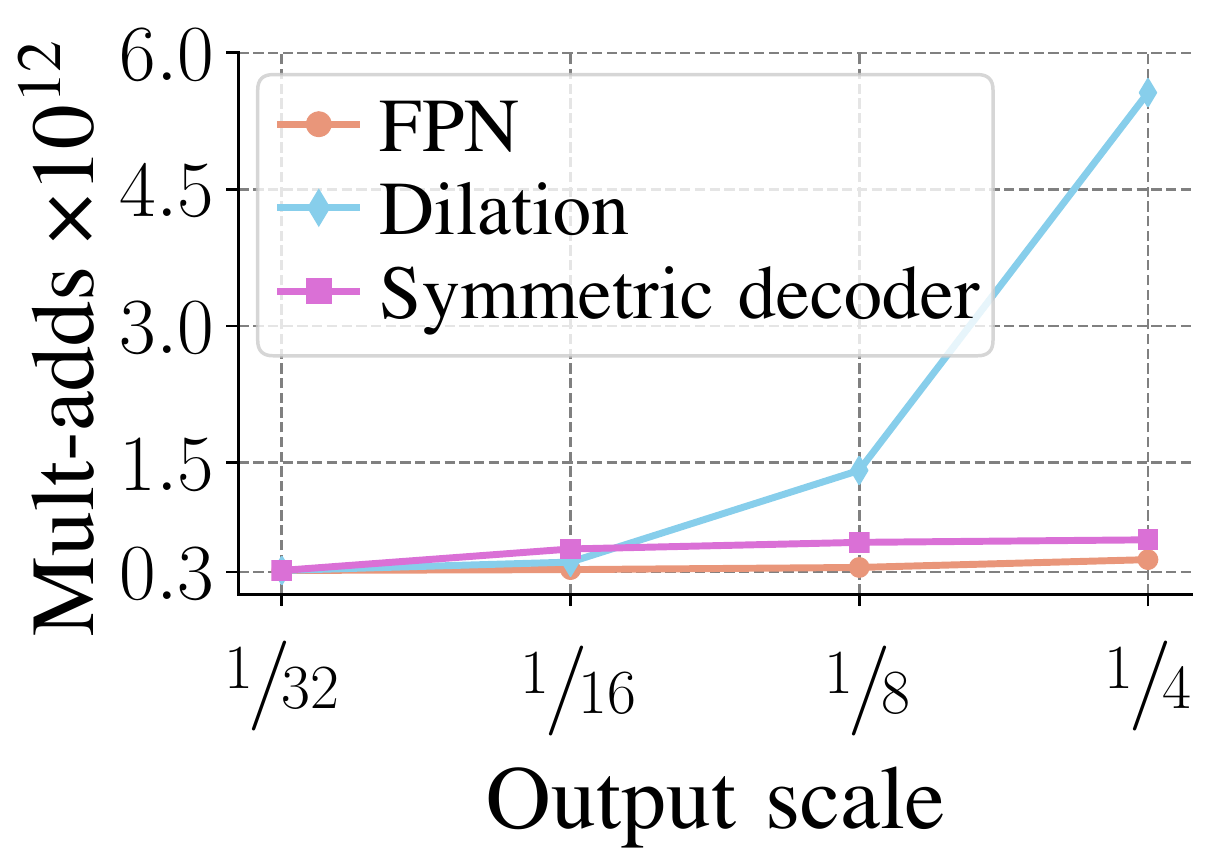}
\includegraphics[width=.495\linewidth]{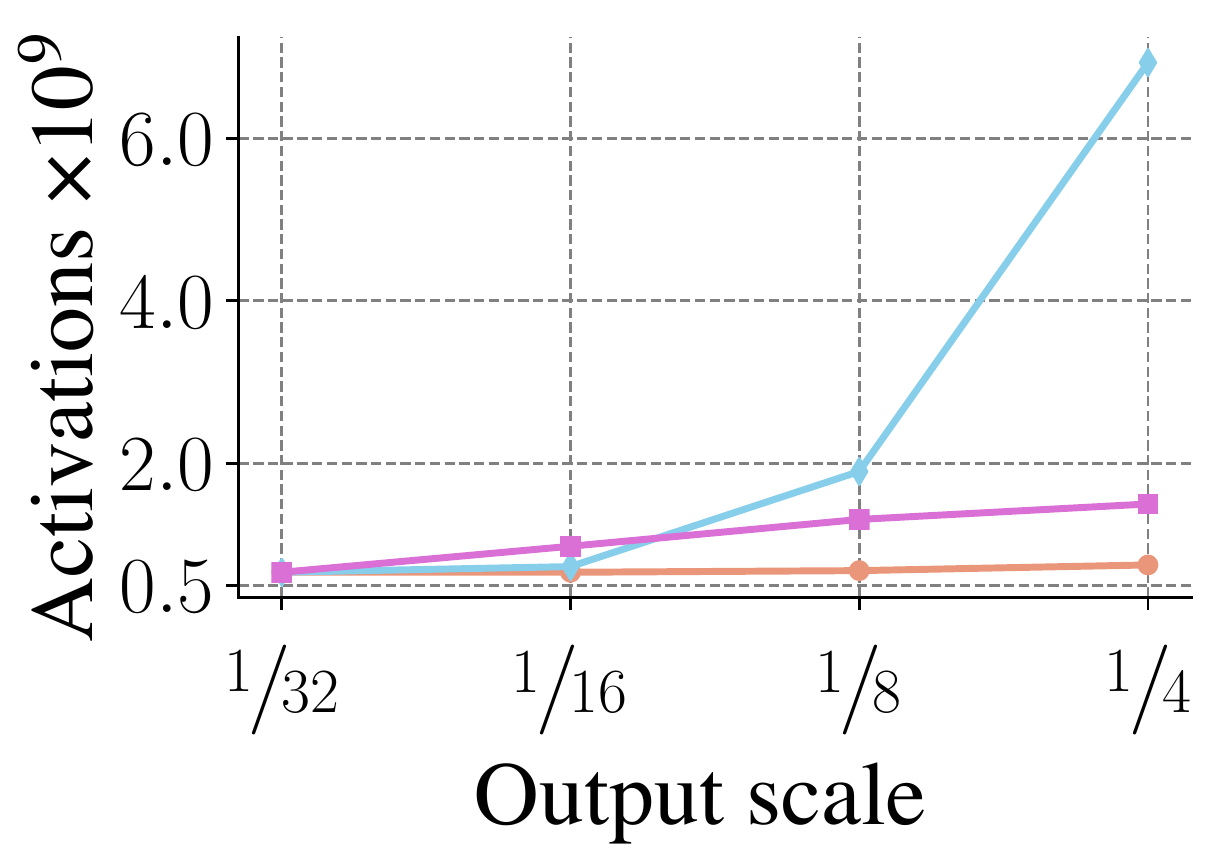}
\caption{\textbf{Backbone architecture efficiency.} We compare methods for increasing feature resolution for semantic segmentation, including dilated networks, symmetric decoders, and FPN, see Figure \ref{fig:backbone}. We count multiply-adds and memory used when applying ResNet-101 to a 2 megapixel image. FPN at output scale 1/4 is similar computationally to dilation-16 (1/16 resolution output), but produces a 4\x~higher resolution output. Increasing resolution to 1/8 via dilation uses a further \app3\x~more compute and memory.}
\label{fig:efficiency}
\end{figure}
%##############################################################################################

%##############################################################################################
\begin{figure*}
\includegraphics[width=1.0\linewidth]{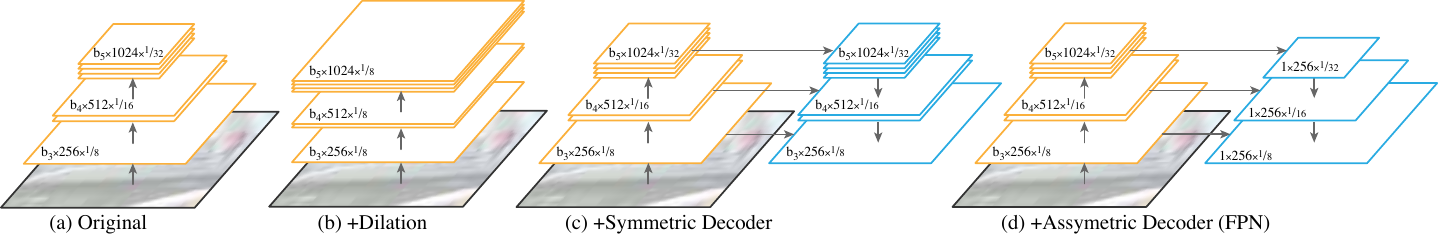}
\caption{\textbf{Backbone architectures} for increasing feature resolution. (a) A standard convolutional network (dimensions are denoted as \#blocks\x\#channels\x resolution). (b) A common approach is to reduce the stride of select convolutions and use dilated convolutions after to compensate. (c) A U-Net~\cite{ronneberger2015u} style network uses a \emph{symmetric} decoder that mirrors the bottom-up pathway, but in reverse. (d) FPN can be seen as an \emph{asymmetric}, lightweight decoder whose top-down pathway has only one block per stage and uses a shared channel dimension. For a comparison of the efficiency of these models, please see Figure~\ref{fig:efficiency}.}
\label{fig:backbone}
\end{figure*}
%##############################################################################################

\paragraph{Joint training:} During training the instance segmentation branch has three losses~\cite{he2017mask}: $L_\texttt{c}$ (classification loss), $L_\texttt{b}$ (bounding-box loss), and $L_\texttt{m}$ (mask loss). The total instance segmentation loss is the sum of these losses, where $L_\texttt{c}$ and $L_\texttt{b}$ are normalized by the number of sampled  RoIs and $L_\texttt{m}$ is normalized by the number of foreground RoIs. The semantic segmentation loss, $L_\texttt{s}$, is computed as a per-pixel cross entropy loss between the predicted and the ground-truth labels, normalized by the number of labeled image pixels.

We have observed that the losses from these two branches have different scales and normalization policies. Simply adding them \emph{degrades} the final performance for \emph{one} of the tasks. This can be corrected by a simple loss re-weighting between the total instance segmentation loss and the semantic segmentation loss. Our final loss is thus: $L = \lambda_\texttt{i} \left( L_\texttt{c} +  L_\texttt{b} + L_\texttt{m} \right) + \lambda_\texttt{s} L_\texttt{s}$. By tuning $\lambda_\texttt{i}$ and $\lambda_\texttt{s}$ it is possible to train a single model that is comparable to two separate task-specific models, but at about half the compute.

\subsection{Analysis}

Our motivation for predicting semantic segmentation using FPN is to create a simple, single-network baseline that can perform both instance and semantic segmentation. However, it is also interesting to consider the memory and computational footprint of our approach relative to model architectures popular for semantic segmentation. The most common designs that produce high-resolution outputs are dilated convolution (Figure~\ref{fig:backbone}b) and \emph{symmetric} encoder-decoder models that have a mirror image decoder with lateral connections (Figure~\ref{fig:backbone}c). While our primary motivation is compatibility with Mask R-CNN, we note that FPN is much lighter than a typically used dilation-8 network, \app2\x~more efficient than the symmetric encoder-decoder, and roughly equivalent to a dilation-16 network (while producing a 4\x~higher resolution output). See Figure~\ref{fig:efficiency}.

%%%%%%%%%%%%%%%%%%%%%%%%%%%%%%%%%%%%%%%%%%%%%%%%%%%%%%%%%%%%%%%%%%%%%%%%%%%%%%%%%%%%%%%%%%%%%%%%%%%
\section{Experiments}

Our goal is to demonstrate that our approach, Panoptic FPN, can serve as a simple and effective \emph{single-network baseline} for instance segmentation, semantic segmentation, and their joint task of panoptic segmentation~\cite{kirillov2017panoptic}. For instance segmentation, this is expected, since our approach extends Mask R-CNN with FPN. For semantic segmentation, as we simply attach a lightweight dense-pixel prediction branch (Figure~\ref{fig:semantic-branch}) to FPN, we need to demonstrate it can be competitive with recent methods. Finally, we must show that Panoptic FPN can be trained in a multi-task setting without loss in accuracy on the individual tasks.

\begin{figure*}
\includegraphics[height=2.263cm]{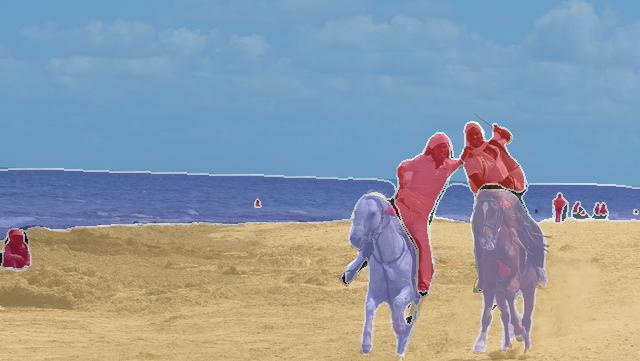}
\includegraphics[height=2.263cm]{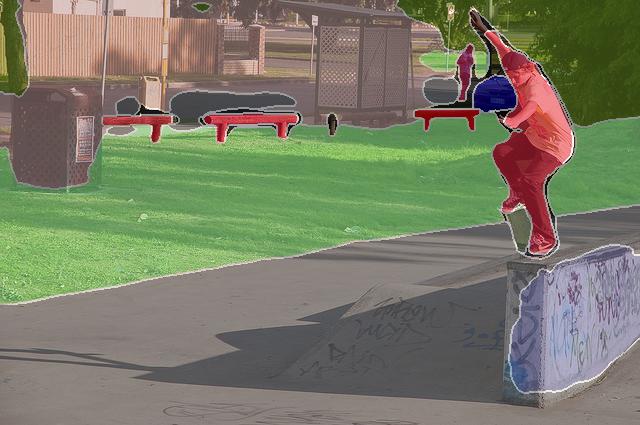}
\includegraphics[height=2.263cm]{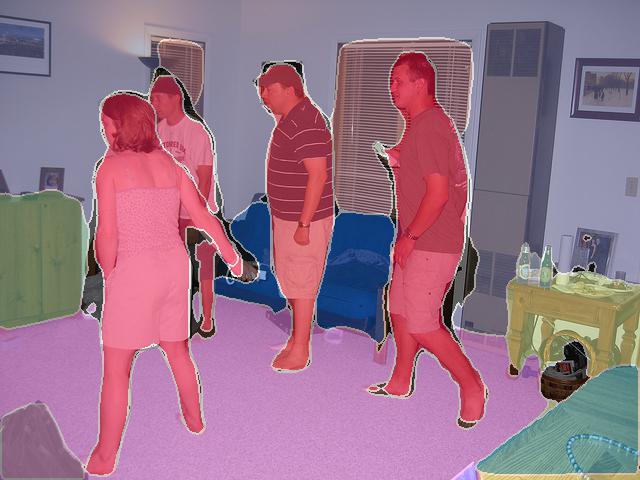}
\includegraphics[height=2.263cm]{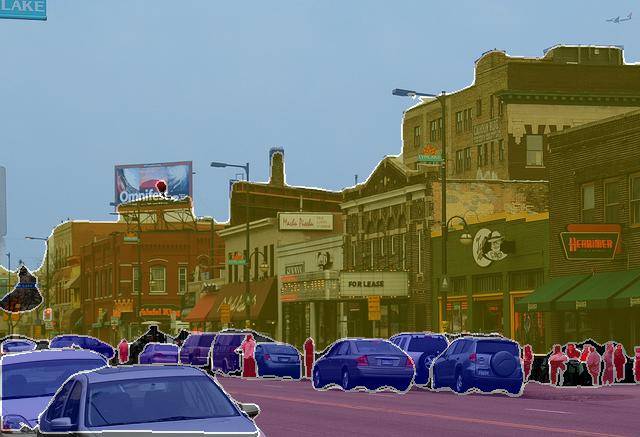}
\includegraphics[height=2.263cm]{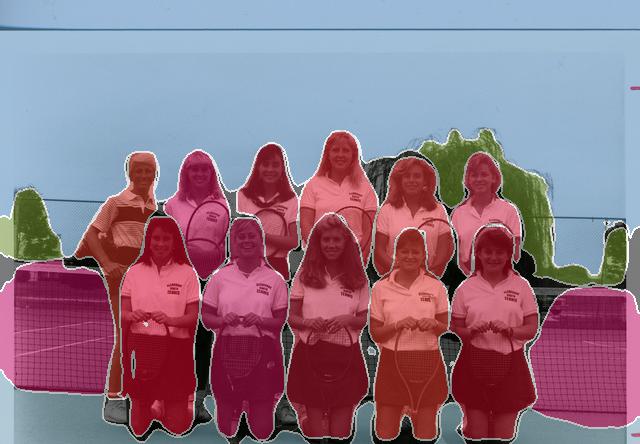}\\[0.5mm]
\includegraphics[width=0.245\linewidth]{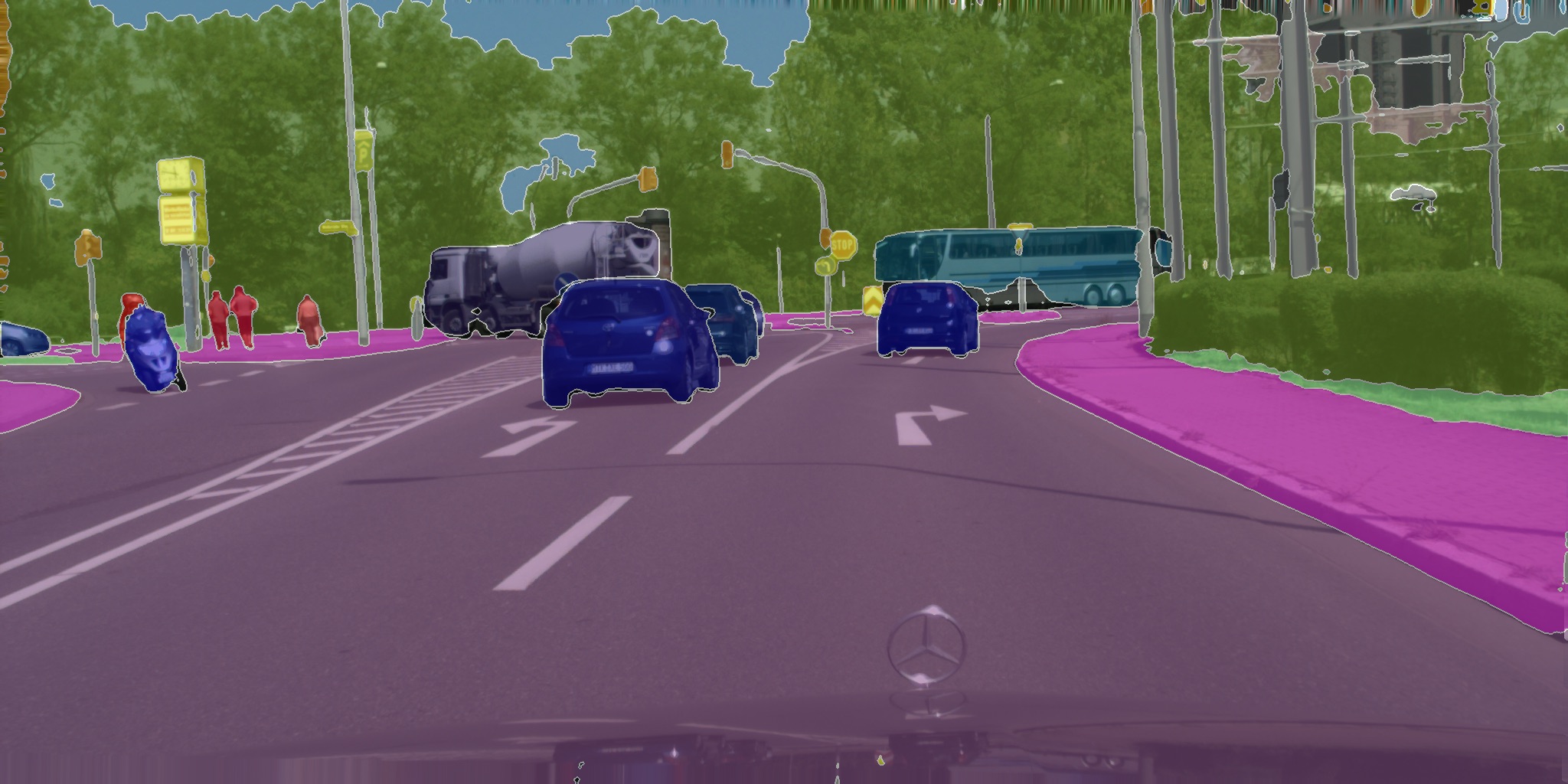}
\includegraphics[width=0.245\linewidth]{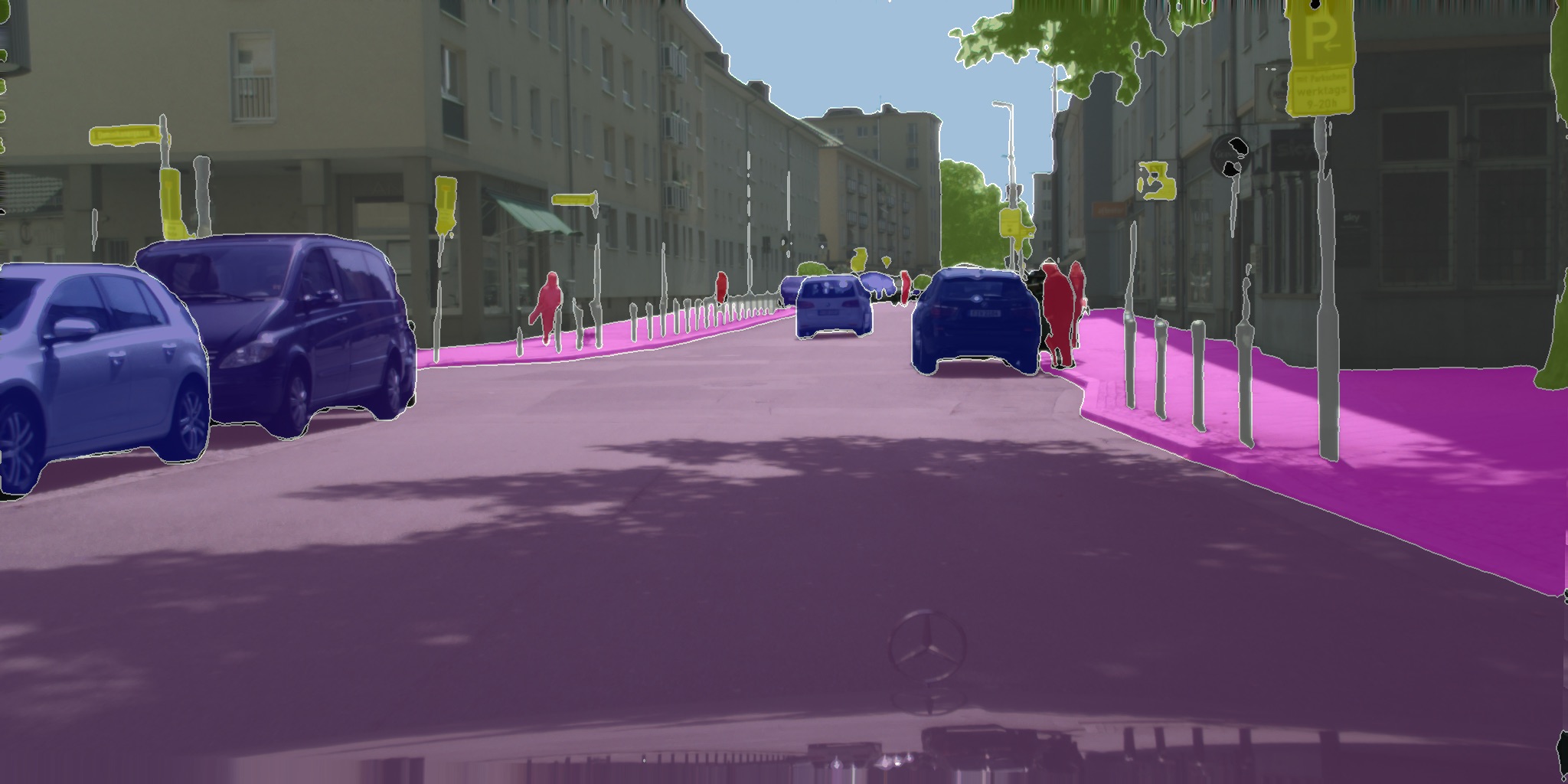}
\includegraphics[width=0.245\linewidth]{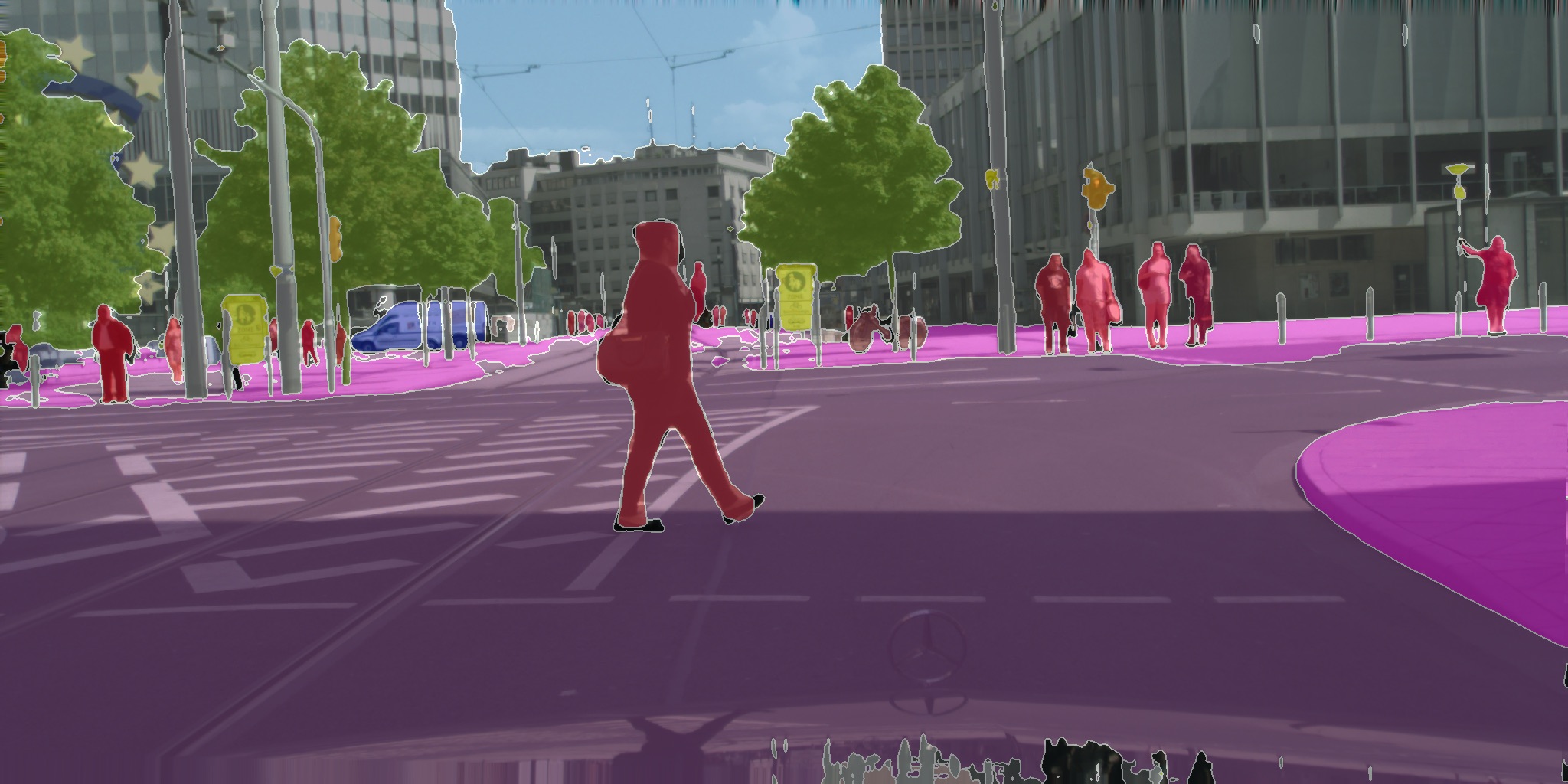}
\includegraphics[width=0.245\linewidth]{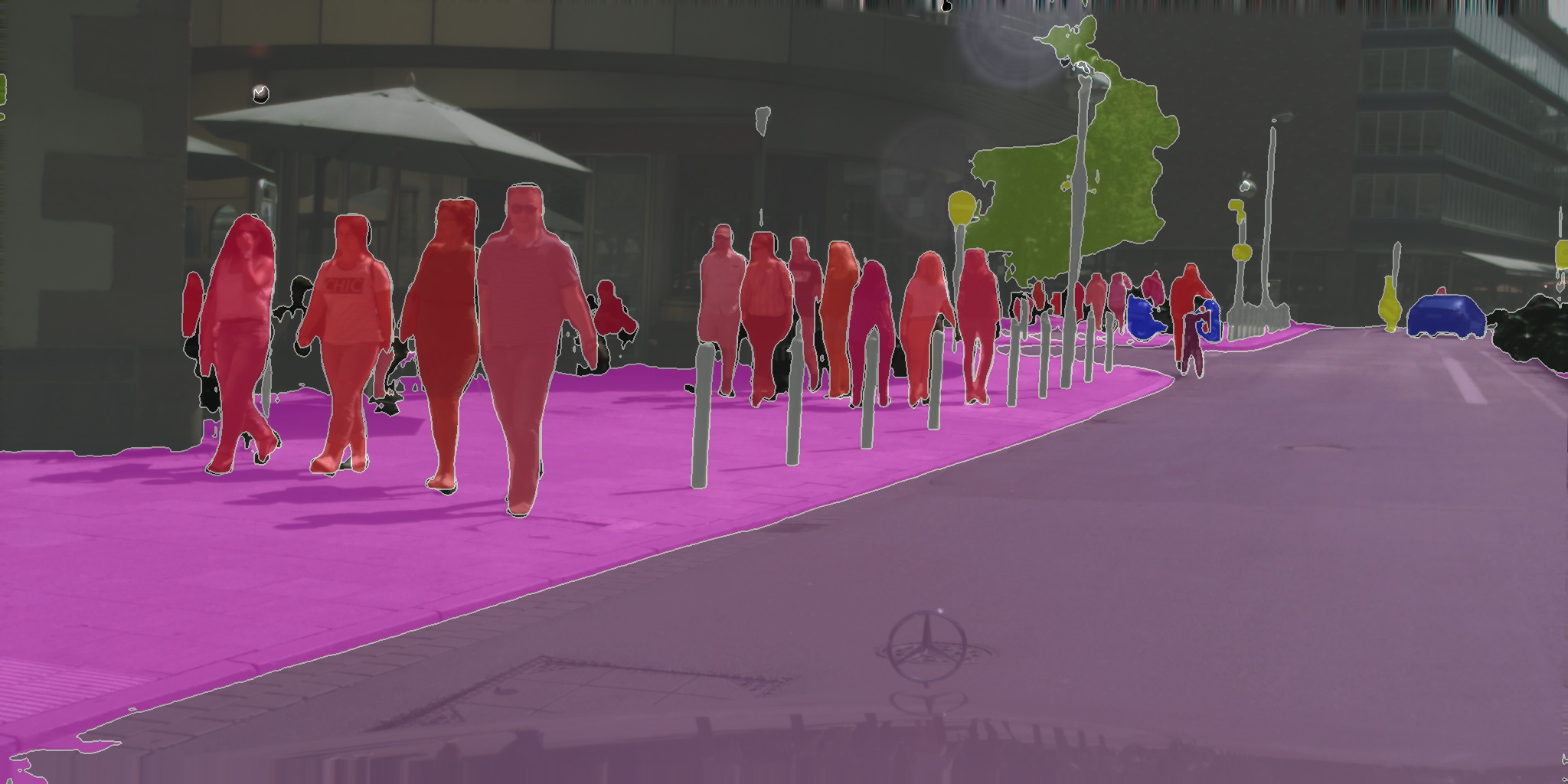}\vspace{-2mm}
\caption{More Panoptic FPN results on COCO (top) and Cityscapes (bottom) using a single ResNet-101-FPN network.}
\label{fig:results:2}\vspace{-2mm}
\end{figure*}

We therefore begin our analysis by testing our approach for semantic segmentation (we refer to this single-task variant as \emph{Semantic FPN}). Surprisingly, this simple model achieves competitive semantic segmentation results on the COCO~\cite{lin2014coco} and Cityscapes~\cite{Cordts2016Cityscapes} datasets. Next, we analyze the integration of the semantic segmentation branch with Mask R-CNN, and the effects of joint training. Lastly, we show results for panoptic segmentation, again on COCO and Cityscapes. Qualitative results are shown in Figures \ref{fig:results:1} and \ref{fig:results:2}. We describe the experimental setup next.

\subsection{Experimental Setup}

\paragraph{COCO:} The COCO dataset~\cite{lin2014coco} was developed with a focus on instance segmentation, but more recently stuff annotations were added~\cite{caesar2016coco}. For instance segmentation, we use the 2017 data splits with 118k/5k/20k train/val/test images and 80 thing classes. For semantic segmentation, we use the 2017 stuff data with 40k/5k/5k splits and 92 stuff classes. Finally, panoptic segmentation~\cite{kirillov2017panoptic} uses all 2017 COCO images with 80 thing and 53 stuff classes annotated.

\paragraph{Cityscapes:} Cityscapes~\cite{Cordts2016Cityscapes} is an ego-centric street-scene dataset. It has 5k high-resolution images (1024\x2048 pixels) with fine pixel-accurate annotations: 2975 train, 500 val, and 1525 test. An additional 20k images with coarse annotations are available, we do \emph{not} use these in our experiments. There are 19 classes, 8 with instance-level masks.

\paragraph{Single-task metrics:} We report standard semantic and instance segmentation metrics for the individual tasks using evaluation code provided by each dataset. For semantic segmentation, the \textbf{mIoU} (mean Intersection-over-Union)~\cite{everingham2015pascal} is the primary metric on both COCO and Cityscapes. We also report fIoU (frequency weighted IoU) on COCO~\cite{caesar2016coco} and iIoU (instance-level IoU) on Cityscapes~\cite{Cordts2016Cityscapes}. For instance segmentation, \textbf{AP} (average precision averaged over categories and IoU thresholds)~\cite{lin2014coco} is the primary metric and AP$_{50}$ and AP$_{75}$ are selected supplementary metrics.

\paragraph{Panoptic segmentation metrics:} We use \textbf{PQ} (panoptic quality) as the default metric to measure Panoptic FPN performance, for details see~\cite{kirillov2017panoptic}. PQ captures both recognition and segmentation quality, and treats both stuff and thing categories in a unified manner. This single, unified metric allows us to directly compare methods. Additionally, we use PQ\stuff and PQ\things to report stuff and thing performance separately. Note that PQ is used to evaluate Panoptic FPN predictions after the post-processing merging procedure is applied to the outputs of the semantic and instance branches.

\paragraph{COCO training:} We use the default Mask R-CNN 1$\times$ training setting~\cite{Detectron2018} with scale jitter (shorter image side in [640, 800]). For semantic segmentation, we predict 53 stuff classes plus a single `other' class for all 80 thing classes.

\paragraph{Cityscapes training:} We construct each minibatch from 32 random 512\x1024 image crops (4 crops per GPU) after randomly scaling each image by 0.5 to 2.0\x. We train for 65k iterations starting with a learning rate of 0.01 and dropping it by a factor of 10 at 40k and 55k iterations. This differs from the original Mask R-CNN setup~\cite{he2017mask} but is effective for both instance and semantic segmentation. For the largest backbones for semantic segmentation, we perform color augmentation~\cite{liu2016ssd} and crop bootstrapping~\cite{bulo2017place}. For semantic segmentation, predicting all thing classes, rather than a single `other' label, performs better (for panoptic inference we discard these predictions). Due to the high variance of the mIoU (up to 0.4), we report the median performance of 5 trials of each experiment on Cityscapes.

%##################################################################################################
\begin{table}[t]\centering
\subfloat[\bd{Cityscapes Semantic FPN.} Performance is reported on the \emph{val} set and all methods use only fine Cityscapes annotations for training. The backbone notation includes the dilated resolution `D' (note that \cite{deeplabV3plus} uses both dilation and an encoder-decoder backbone). All top-performing methods other than ours use dilation. FLOPs (multiply-adds \x$10^{12}$) and memory (\#~activations \x$10^9$) are approximate but informative. For these larger FPN models we train with color and crop augmentation. Our baseline is comparable to state-of-the-art methods in accuracy and efficiency.
\label{tab:separate:semantic:cityscapes}]{
\tablestyle{5pt}{1.0}\begin{tabular}{l|c|ccc}
 & backbone & mIoU & \footnotesize FLOPs & \footnotesize memory \\\shline
 \footnotesize DeeplabV3~\cite{deeplabV3} & \scriptsize ResNet-101-D8
  & 77.8 & 1.9 & 1.9\\
 \footnotesize PSANet101~\cite{zhao2018psanet} & \scriptsize ResNet-101-D8
  & 77.9 & 2.0 & 2.0\\
 \footnotesize Mapillary~\cite{bulo2017place} & \scriptsize WideResNet-38-D8
  & 79.4 & 4.3 & 1.7\\
 \footnotesize DeeplabV3+~\cite{deeplabV3plus} & \scriptsize X-71-D16
  & 79.6 & 0.5 & 1.9\\
\hline
 \footnotesize \bd{Semantic FPN} & \scriptsize ResNet-101-FPN
  & 77.7 & 0.5 & 0.8\\
 \footnotesize \bd{Semantic FPN} & \scriptsize ResNeXt-101-FPN
  & 79.1 & 0.8 & 1.4
\end{tabular}}\\\vspace{-1mm}
% subfloat b
\subfloat[\bd{COCO-Stuff 2017 Challenge results.} We submitted an early version of Semantic FPN to the 2017 COCO Stuff Segmentation Challenge held at ECCV (\url{http://cocodataset.org/\#stuff-2017}). \emph{Our entry won first place} without ensembling, and we outperformed competing methods by at least a 2 point margin on all reported metrics.\label{tab:separate:semantic:coco}]{
\tablestyle{9pt}{1.0}\begin{tabular}{@{}l|c|ccc@{}}
  & backbone & mIoU & fIoU\\\shline
 \footnotesize Vllab~\cite{vllab_stuff} & \scriptsize Stacked Hourglass
  & 12.4 & 38.8\\
 \footnotesize DeepLab VGG16~\cite{deeplabV2} & \scriptsize VGG-16
  & 20.2 & 47.5\\
 \footnotesize Oxford~\cite{oavl_stuff} & \scriptsize ResNeXt-101
  & 24.1 & 50.6\\
 \footnotesize G-RMI~\cite{grmi_stuff} & \scriptsize Inception ResNet v2
  & 26.6 & 51.9\\
\hline
 \footnotesize \bd{Semantic FPN} & \scriptsize ResNeXt-152-FPN
  &  28.8 & 55.7
\end{tabular}}\\\vspace{-1mm}
% subfloat c
\subfloat[\bd{Ablation} (mIoU): Channel width of 128 for the features in the semantic branch strikes a good balance between accuracy and efficiency.\label{tab:separate:semantic:width}]{
\tablestyle{4.5pt}{1.0}\begin{tabular}{c|cc}
 \small Width & \small Cityscapes & \small COCO \\
\shline
 64   & 74.1 & 39.6 \\
 128  & 74.5 & 40.2 \\
 256  & 74.6 & 40.1
\end{tabular}}\hspace{2mm}
% subfloat d
\subfloat[\bd{Ablation} (mIoU): Sum aggregation of the feature maps in the semantic branch is marginally better and is more efficient.\label{tab:separate:semantic:aggregation}]{
\tablestyle{3pt}{1.0}\begin{tabular}{c|cc}
 Aggr. & Cityscapes & COCO \\
\shline
 Sum     & 74.5 & 40.2 \\
 Concat  & 74.4 & 39.9 \\
 \multicolumn{2}{c}{~}
\end{tabular}}\\\vspace{-1mm}
% main caption
\caption{\bd{Semantic Segmentation using FPN.}
\label{tab:panoptic:ablation}}
\end{table}\vspace{-2mm}
%##################################################################################################

%##################################################################################################
\begin{table*}[t]\centering
% subfloat a
\subfloat[Panoptic FPN on COCO for \bd{instance} segmentation ($\lambda_i = 1$).]{
\tablestyle{9.5pt}{1.0}\begin{tabular}{l|c|cccc}
 $\lambda_\texttt{s}$ & mIoU & AP & AP$_{50}$ & AP$_{75}$  & PQ\things \\
\shline
 0.0 & \textcolor{gray}{-}    & 33.9 &	55.6 & 35.9 & 46.6\\
\hline
 0.1 & \textcolor{gray}{37.2} & \bd{34.0} & 55.6 & \bd{36.0} & \bd{46.8} \\
 0.25 & \textcolor{gray}{39.6} & 33.7 & 55.3 & 35.5 & 46.1\\
 0.5 & \textcolor{gray}{41.0} & 33.3 & 54.9 & 35.2 & 45.9\\
 0.75 & \textcolor{gray}{41.1} & 32.6 & 53.9 & 34.6 & 45.0\\
 1.0 & \textcolor{gray}{41.5} & 32.1 & 53.2 & 33.6 & 44.6\\
\hline
 & & \dt{+0.1} & \dt{+0.0} & \dt{+0.1} & \dt{+0.2}
\end{tabular}}\hspace{7mm}
% subfloat b
\subfloat[Panoptic FPN on Cityscapes for \bd{instance} segmentation ($\lambda_i = 1$).]{
\tablestyle{13.5pt}{1.0}\begin{tabular}{l|c|ccc}
 $\lambda_\texttt{s}$ & mIoU & AP & AP$_{50}$ & PQ\things \\
\shline
 0.0 & \textcolor{gray}{-}   & 32.2 & 58.7  & 51.3\\
\hline
 0.1 & \textcolor{gray}{68.3} & 32.5 & 59.2 & 52.9\\
 0.25 & \textcolor{gray}{71.8} & 32.8 & 59.6 & 52.7\\
 0.5 & \textcolor{gray}{72.0} & 32.7 & 59.5 & 52.9\\
 0.75 & \textcolor{gray}{73.4} & 32.8 & 58.8 & 52.3\\
 1.0 & \textcolor{gray}{74.2} & \bd{33.2} & \bd{59.7} & \bd{52.4}\\
\hline
 & & \dt{+1.0} & \dt{+1.0} & \dt{+1.1}
\end{tabular}}\\
% subfloat c
\subfloat[Panoptic FPN on COCO for \bd{semantic} segmentation ($\lambda_s = 1$).]{
\tablestyle{13.5pt}{1.0}\begin{tabular}{l|c|ccc}
 $\lambda_\texttt{i}$ & AP & mIoU & fIoU & PQ\stuff \\
\shline
 0.0 & \textcolor{gray}{-}    & 40.2 & 67.2 & 27.9\\
\hline
 0.1 & \textcolor{gray}{20.1} & 40.6 & 67.5 & 28.4\\
 0.25 & \textcolor{gray}{25.5} & 41.0 & 67.8 & 28.6\\
 0.5 & \textcolor{gray}{29.2} & 41.3 & 68.0 & 28.9\\
 0.75 & \textcolor{gray}{30.8} & 41.1 & 68.2 & 28.9\\
 1.0 & \textcolor{gray}{32.1} & \bd{41.5} & \bd{68.2} & \bd{29.0}\\
\hline
 & & \dt{+1.2} & \dt{+1.0} & \dt{+1.1}
\end{tabular}}\hspace{7mm}
% subfloat d
\subfloat[Panoptic FPN on Cityscapes for \bd{semantic} segmentation ($\lambda_s = 1$).]{
\tablestyle{13.5pt}{1.0}\begin{tabular}{l|c|ccc}
 $\lambda_\texttt{i}$ & AP & mIoU & iIoU & PQ\stuff \\
\shline
 0.0 & \textcolor{gray}{-}    & 74.5 & 55.8 & 62.4\\
\hline
 0.1 & \textcolor{gray}{27.4} & 75.3 & 57.6 & 62.5\\
 0.25 & \textcolor{gray}{30.5} & \bd{75.5} & \bd{58.3} & \bd{62.5}\\
 0.5 & \textcolor{gray}{32.0} & 75.0 & 58.2 & 62.2\\
 0.75 & \textcolor{gray}{32.6} & 74.3 & 58.2 & 61.7\\
 1.0 & \textcolor{gray}{33.2} & 74.2 & 57.4 & 61.4\\
\hline
 & & \dt{+1.0} & \dt{+2.5} & \dt{+0.1}
\end{tabular}}
% main caption
\caption{\textbf{Multi-Task Training:} (a,b) \emph{Adding a semantic segmentation branch can slightly improve instance segmentation results} over a single-task baseline with properly tuned $\lambda_\texttt{s}$ (results bolded). Note that $\lambda_\texttt{s}$ indicates the weight assigned to the semantic segmentation loss and $\lambda_\texttt{s}=0.0$ serves as the single-task baseline. (c,d) \emph{Adding an instance segmentation branch can provide even stronger benefits for semantic segmentation} over a single-task baseline with properly tuned $\lambda_\texttt{i}$ (results bolded). As before, $\lambda_\texttt{i}$ indicates the weight assigned to the instance segmentation loss and $\lambda_\texttt{i}=0.0$ serves as the single-task baseline. While promising, we are more interested in the joint task, for which results are shown in Table \ref{tab:panoptic}.}
\label{tab:multitask}\vspace{3mm}
\end{table*}
%##################################################################################################

%##################################################################################################
\begin{table*}[t]\centering
% subfloat a
\subfloat[\bd{Panoptic Segmentation: Panoptic \emph{R50-FPN} \vs R50-FPN\x2}. \emph{Using a single FPN network for solving both tasks simultaneously yields comparable accuracy to two independent FPN networks for instance and semantic segmentation, but with roughly half the compute}.\label{tab:panoptic:RN50}]{
\tablestyle{5pt}{1.0}\begin{tabular}{l|l|cc|cc|c}
 & backbone & AP & PQ\things & mIoU & PQ\stuff & PQ\\
\shline
 \multirow{3}{*}{\scriptsize COCO}
 & \scriptsize R50-FPN\x2 & 33.9 & 46.6 & 40.2 & 27.9 & 39.2 \\
 & \scriptsize R50-FPN    & 33.3 & 45.9 & 41.0 & 28.7 & 39.0 \\
 & & \dt{-0.6} & \dt{-0.7} & \dt{+0.8} & \dt{+0.8} & \dt{-0.2}\\
\hline
 \multirow{3}{*}{\scriptsize Cityscapes}
 & \scriptsize R50-FPN\x2 & 32.2 & 51.3 & 74.5 & 62.4 & 57.7\\
 & \scriptsize R50-FPN    & 32.0 & 51.6 & 75.0 & 62.2 & 57.7\\
 & & \dt{-0.2} & \dt{+0.3} & \dt{+0.5} & \dt{-0.2} & \dt{+0.0}\\
\end{tabular}}\hspace{7mm}
% subfloat b
\subfloat[\bd{Panoptic Segmentation: Panoptic \emph{R101-FPN} \vs R50-FPN\x2}. \emph{Given a roughly \emph{equal} computational budget, a single FPN network for the panoptic task outperforms two independent FPN networks for instance and semantic segmentation by a healthy margin.}\label{tab:panoptic:RN101}]{
\tablestyle{5pt}{1.0}\begin{tabular}{l|l|cc|cc|c}
 & backbone & AP & PQ\things & mIoU & PQ\stuff & PQ\\
\shline
 \multirow{3}{*}{\scriptsize COCO}
 & \scriptsize R50-FPN\x2 & 33.9 & 46.6 & 40.2 & 27.9 & 39.2\\
 & \scriptsize R101-FPN   & 35.2 & 47.5 & 42.1 & 29.5 & 40.3\\
 & & \dt{+1.3} & \dt{+0.9} & \dt{+1.9} & \dt{+1.6} & \dt{+1.1}\\
\hline
 \multirow{3}{*}{\scriptsize Cityscapes}
 & \scriptsize R50-FPN\x2 & 32.2 & 51.3 & 74.5 & 62.4 & 57.7 \\
 & \scriptsize R101-FPN   & 33.0 & 52.0 & 75.7 & 62.5 & 58.1 \\
 & & \dt{+0.8} & \dt{+0.7} & \dt{+1.3} & \dt{+0.1} & \dt{+0.4}\\
\end{tabular}}\\
% subfloat c
\subfloat[\bd{Training Panoptic FPN}. During training, for each minibatch we can either \emph{combine} the semantic and instances loss or we can \emph{alternate} which loss we compute (in the latter case we train for twice as long). We find that combining the losses in each minibatch performs much better.\label{tab:panoptic:interleave}]{
\tablestyle{6pt}{1.0}\begin{tabular}{l|l|cc|cc|c}
 & loss & AP & PQ\things & mIoU & PQ\stuff & PQ\\
\shline
 \multirow{3}{*}{\scriptsize COCO}
 & \scriptsize alternate & 31.7 & 43.9 & 40.2 & 28.0 & 37.5\\
 & \scriptsize combine   & 33.3 & 45.9 & 41.0 & 28.7 & 39.0\\
 & & \dt{+1.6} & \dt{+2.0} & \dt{+0.8} & \dt{+0.7} & \dt{+1.5}\\
\hline
 \multirow{3}{*}{\scriptsize Cityscapes}
 & \scriptsize alternate  & 32.0 & 51.4 & 74.3 & 61.3 & 57.4\\
 & \scriptsize combine    & 32.0 & 51.6 & 75.0 & 62.2 & 57.7\\
 & & \dt{+0.0} & \dt{+0.2} & \dt{+0.7} & \dt{+0.9} & \dt{+0.3}\\
\end{tabular}}\hspace{7mm}
% subfloat d
\subfloat[\bd{Grouped FPN}. We test a variant of Panoptic FPN where we group the 256 FPN channels into two sets and apply the instance and semantic branch to its own dedicated group of 128. While this gives mixed gains, we expect better multi-task strategies can improve results.
\label{tab:panoptic:split}]{
\tablestyle{6pt}{1.0}\begin{tabular}{l|c|cc|cc|c}
 & FPN & AP & PQ\things & mIoU & PQ\stuff & PQ\\
\shline
 \multirow{3}{*}{\scriptsize COCO}
 & \scriptsize original & 33.3 & 45.9 & 41.0 & 28.7 & 39.0\\
 & \scriptsize grouped  & 33.1 & 45.7 & 41.2 & 28.4 & 38.8\\
 & & \dt{-0.2} & \dt{-0.2} & \dt{+0.2} & \dt{-0.3} & \dt{-0.2}\\
\hline
 \multirow{3}{*}{\scriptsize Cityscapes}
 & \scriptsize original & 32.0 & 51.6 & 75.0 & 62.2 & 57.7\\
 & \scriptsize grouped  & 32.0 & 51.8 & 75.3 & 61.7 & 57.5\\
 & & \dt{+0.0} & \dt{+0.2} & \dt{+0.3} & \dt{-0.5} & \dt{-0.2}\\
\end{tabular}}\\
% main caption
\caption{\textbf{Panoptic FPN Results.}}
\label{tab:panoptic}
\end{table*}
%##################################################################################################

\subsection{FPN for Semantic Segmentation}

\paragraph{Cityscapes:} We start by comparing our \emph{baseline} Semantic FPN to existing methods on the Cityscapes val split in Table~\ref{tab:separate:semantic:cityscapes}. We compare to recent top-performing methods, but not to \emph{competition entires} which typically use ensembling, COCO pre-training, test-time augmentation, \etc. Our approach, which is a minimal extension to FPN, is able to achieve strong results compared to systems like DeepLabV3+~\cite{deeplabV3plus}, which have undergone many design iterations. In terms of compute and memory, Semantic FPN is lighter than typical dilation models, while yielding higher resolution features (see Fig.~\ref{fig:efficiency}). We note that adding dilation into FPN could potentially yield further improvement but is outside the scope of this work. Moreover, in our baseline we deliberately avoid orthogonal architecture improvements like Non-local~\cite{wang2018non} or SE~\cite{hu2017squeeze}, which would likely yield further gains. Overall, these results demonstrate that our approach is a strong baseline for semantic segmentation.

\paragraph{COCO:} An earlier version of \emph{our approach won the 2017 COCO-Stuff challenge}. Results are reported in Table~\ref{tab:separate:semantic:coco}. As this was an early design, the the semantic branch differed slightly (each upsampling module had two 3\x3 conv layers and ReLU before bilinear upscaling to the final resolution, and features were concatenated instead of summed, please compare with Figure~\ref{fig:semantic-branch}). As we will show in the ablations shortly, results are fairly robust to the exact branch design. Our competition entry was trained with color augmentation~\cite{liu2016ssd} and at test time balanced the class distribution and used multi-scale inference. Finally, we note that at the time we used a training schedule specific to semantic segmentation similar to our Cityscapes schedule (but with double learning rate and halved batch size).

\paragraph{Ablations:} We perform a few ablations to analyze our proposed semantic segmentation branch (shown in Figure~\ref{fig:semantic-branch}). For consistency with further experiments in our paper, we use stuff annotations from the COCO Panoptic dataset (which as discussed differ from those used for the COCO Stuff competition). Table~\ref{tab:separate:semantic:width} shows ResNet-50 Semantic FPN with varying number of channels in the semantic branch. We found that 128 strikes a good balance between accuracy and efficiency. In Table~\ref{tab:separate:semantic:aggregation} we compare element-wise sum and concatenation for aggregating feature maps from different FPN levels. While accuracy for both is comparable, summation is more efficient. Overall we observe that the simple architecture of the new dense-pixel labelling branch is robust to exact design choices.

\subsection{Multi-Task Training}

Single-task performance of our approach is quite effective; for semantic segmentation the results in the previous section demonstrate this, for instance segmentation this is known as we start from Mask R-CNN. However, can we jointly train for both tasks in a multi-task setting?

To combine our semantic segmentation branch with the instance segmentation branch in Mask R-CNN, we need to determine how to train a single, unified network. Previous work demonstrates that multi-task training is often challenging and can lead to degraded results~\cite{kokkinos2016ubernet, kendall2017multi}. We likewise observe that for semantic or instance segmentation, adding the secondary task can degrade the accuracy in comparison with the single-task baseline.

In Table~\ref{tab:multitask} we show that with ResNet-50-FPN, using a simple loss scaling weight on the semantic segmentation loss, $\lambda_\texttt{s}$, or instance segmentation loss, $\lambda_\texttt{i}$, we can obtain a re-weighting that improves results over single-task baselines. Specifically, adding a semantic segmentation branch with the proper $\lambda_\texttt{s}$ improves instance segmentation, and vice-versa. This can be exploited to improve single-task results. However, our main goal is to solve both tasks simultaneously, which we explore in the next section.

%##############################################################################################
%##############################################################################################

%##################################################################################################
\begin{table*}[t]\centering
% subfloat a
\subfloat[\bd{Panoptic Segmentation on COCO test-dev.} We submit Panoptic FPN to the COCO test-dev leaderboard (for details on competing entries, please see \url{http://cocodataset.org/\#panoptic-leaderboard}). We only compare to entires that use a \emph{single network} for the joint task. We do \emph{not} compare to competition-level entires that utilize ensembling (including methods that ensemble separate networks for semantic and instance segmentation). For methods that use \emph{one network} for panoptic segmentation, our approach improves PQ by an \app9 point margin.
\label{tab:soa:coco}]{
\tablestyle{11.9pt}{1.0}\begin{tabular}{l|ccc}
 & PQ & PQ\things & PQ\stuff\\
\shline
 Artemis & 16.9 & 16.8 & 17.0\\
 LeChen & 26.2 & 31.0 & 18.9\\
 MPS-TU Eindhoven~\cite{de2018panoptic} & 27.2 & 29.6 & 23.4\\
 MMAP-seg & 32.1 & 38.9 & 22.0\\
\hline
 Panoptic FPN & \bd{40.9} & \bd{48.3} & \bd{29.7}\\
\end{tabular}}\hspace{5mm}\vspace{-1mm}
% subfloat b
\subfloat[\bd{Panoptic Segmentation on Cityscapes.} For Cityscapes, there is no public leaderboard for panoptic segmentation at this time. Instead, we compare on val to the recent work of Arnab and Torr~\cite{arnab2017pixelwise,li2018weakly} who develop a novel approach for panoptic segmentation, named DIN. DIN is representative of alternatives to region-based instance segmentation that start with a pixel-wise semantic segmentation and then perform grouping to extract instances (see the related work). Panoptic FPN, without extra coarse training data or any bells and whistles, outperforms DIN by a 4.3 point PQ margin.
\label{tab:soa:cs}]{
\tablestyle{5.5pt}{1.0}\begin{tabular}{l|c|ccc|cc}
 & \footnotesize{coarse} & PQ & PQ\things & PQ\stuff & mIoU & AP \\
\shline
 DIN~\cite{arnab2017pixelwise,li2018weakly} & \checkmark & 53.8 & 42.5 & 62.1 & \bd{80.1} & 28.6 \\
 Panoptic FPN &  & \bd{58.1} & \bd{52.0} & \bd{62.5} & 75.7 & \bd{33.0} \\
 \multicolumn{3}{c}{~}\\ \multicolumn{3}{c}{~}\\ \multicolumn{3}{c}{~}\\
\end{tabular}}\vspace{-1mm}
% main caption
\caption{\bd{Comparisons of ResNet-101 Panoptic FPN to the state of the art.}}
\label{tab:soa}\vspace{-2mm}
\end{table*}
%##################################################################################################

\subsection{Panoptic FPN}
We now turn to our main result: testing Panoptic FPN for the joint task of panoptic segmentation~\cite{kirillov2017panoptic}, where the network must jointly and accurately output stuff and thing segmentations. For the following experiments, for each setting we select the optimal  $\lambda_\texttt{s}$ and $\lambda_\texttt{i}$ from $\{0.5, 0.75, 1.0\}$, ensuring that results are not skewed by fixed choice of $\lambda$'s.

\paragraph{Main results:} In Table~\ref{tab:panoptic:RN50} we compare two networks trained separately to Panoptic FPN with a single backbone. \emph{Panoptic FPN yields comparable accuracy but with roughly half the compute} (the backbone dominates compute, so the reduction is almost 50\%).  We also balance computational budgets by comparing two separate networks with ResNet-50 backbones each and Panoptic FPN with ResNet-101, see Table~\ref{tab:panoptic:RN101}. \emph{Using roughly equal computational budget, Panoptic FPN significantly outperforms two separate networks}. Taken together, these results demonstrate that the joint approach is strictly beneficial, and that our Panoptic FPN can serve as a solid baseline for the joint task.

\paragraph{Ablations:} We perform additional ablations on Panoptic FPN with ResNet-50. First, by default, we combine the instance and semantic losses together during each gradient update. A different strategy is to alternate the losses on each iteration (this may be useful as different augmentation strategies can be used for the two tasks). We compare these two options in Table~\ref{tab:panoptic:interleave}; the combined loss demonstrates better performance. Next, in Table~\ref{tab:panoptic:split} we compare with an architecture where FPN channels are grouped into two sets, and each task uses one of the two features sets as its input. While the results are mixed, we expect more sophisticated multi-task approaches could give stronger gains.

\paragraph{Comparisons:} We conclude by comparing Panoptic FPN with existing methods. For these experiments, we use Panoptic FPN with a ResNet-101 backbone and without bells-and-whistles. In Table~\ref{tab:soa:coco} we show that Panoptic FPN substantially outperforms all \emph{single-model} entries in the recent COCO Panoptic Segmentation Challenge. This establishes a new baseline for the panoptic segmentation task. On Cityscapes, we compare Panoptic FPN with an approach for panoptic segmentation recently proposed in~\cite{arnab2017pixelwise} in Table~\ref{tab:soa:cs}. Panoptic FPN outperforms \cite{arnab2017pixelwise} by a 4.3 point PQ margin.

%%%%%%%%%%%%%%%%%%%%%%%%%%%%%%%%%%%%%%%%%%%%%%%%%%%%%%%%%%%%%%%%%%%%%%%%%%%%%%%%%%%%%%%%%%%%%%%%%%%
\section{Conclusion}

We introduce a conceptually simple yet effective baseline for panoptic segmentation. The method starts with Mask R-CNN with FPN and adds to it a lightweight semantic segmentation branch for dense-pixel prediction. We hope it can serve as a strong foundation for future research.

%%%%%%%%%%%%%%%%%%%%%%%%%%%%%%%%%%%%%%%%%%%%%%%%%%%%%%%%%%%%%%%%%%%%%%%%%%%%%%%%%%%%%%%%%%%%%%%%%%%
{\small \bibliographystyle{ieee} \bibliography{panopticfpn}}

\begin{thebibliography}{10}\itemsep=-1pt

\bibitem{arnab2017pixelwise}
A.~Arnab and P.~H. Torr.
\newblock Pixelwise instance segmentation with a dynamically instantiated
  network.
\newblock In {\em CVPR}, 2017.

\bibitem{badrinarayanan2015segnet}
V.~Badrinarayanan, A.~Kendall, and R.~Cipolla.
\newblock Segnet: A deep convolutional encoder-decoder architecture for image
  segmentation.
\newblock {\em PAMI}, 2017.

\bibitem{bell2016inside}
S.~Bell, C.~Lawrence~Zitnick, K.~Bala, and R.~Girshick.
\newblock Inside-outside net: Detecting objects in context with skip pooling
  and recurrent neural networks.
\newblock In {\em CVPR}, 2016.

\bibitem{oavl_stuff}
P.~Bilinski and V.~Prisacariu.
\newblock {COCO}-{S}tuff 2017 {C}hallenge: {O}xford {A}ctive {V}ision {L}ab
  team.
\newblock 2017.

\bibitem{bulo2017place}
S.~R. Bul{\`o}, L.~Porzi, and P.~Kontschieder.
\newblock In-place activated batchnorm for memory-optimized training of {DNN}s.
\newblock In {\em CVPR}, 2018.

\bibitem{caesar2016coco}
H.~Caesar, J.~Uijlings, and V.~Ferrari.
\newblock {COCO-Stuff}: Thing and stuff classes in context.
\newblock In {\em CVPR}, 2018.

\bibitem{cai2018cascade}
Z.~Cai and N.~Vasconcelos.
\newblock {Cascade R-CNN}: Delving into high quality object detection.
\newblock In {\em CVPR}, 2018.

\bibitem{cao2018triply}
J.~Cao, Y.~Pang, and X.~Li.
\newblock Triply supervised decoder networks for joint detection and
  segmentation.
\newblock {\em arXiv preprint arXiv:1809.09299}, 2018.

\bibitem{chen2017masklab}
L.-C. Chen, A.~Hermans, G.~Papandreou, F.~Schroff, P.~Wang, and H.~Adam.
\newblock {MaskLab}: Instance segmentation by refining object detection with
  semantic and direction features.
\newblock In {\em CVPR}, 2018.

\bibitem{deeplabV2}
L.-C. Chen, G.~Papandreou, I.~Kokkinos, K.~Murphy, and A.~L. Yuille.
\newblock {DeepLab}: Semantic image segmentation with deep convolutional nets,
  atrous convolution, and fully connected {CRF}s.
\newblock {\em PAMI}, 2018.

\bibitem{deeplabV3}
L.-C. Chen, G.~Papandreou, F.~Schroff, and H.~Adam.
\newblock Rethinking atrous convolution for semantic image segmentation.
\newblock {\em arXiv:1706.05587}, 2017.

\bibitem{deeplabV3plus}
L.-C. Chen, Y.~Zhu, G.~Papandreou, F.~Schroff, and H.~Adam.
\newblock Encoder-decoder with atrous separable convolution for semantic image
  segmentation.
\newblock In {\em ECCV}, 2018.

\bibitem{vllab_stuff}
J.-T. Chien and H.-T. Chen.
\newblock {COCO}-{S}tuff 2017 {C}hallenge: {V}llab team.
\newblock 2017.

\bibitem{Cordts2016Cityscapes}
M.~Cordts, M.~Omran, S.~Ramos, T.~Rehfeld, M.~Enzweiler, R.~Benenson,
  U.~Franke, S.~Roth, and B.~Schiele.
\newblock The cityscapes dataset for semantic urban scene understanding.
\newblock In {\em CVPR}, 2016.

\bibitem{dai2017deformable}
J.~Dai, H.~Qi, Y.~Xiong, Y.~Li, G.~Zhang, H.~Hu, and Y.~Wei.
\newblock Deformable convolutional networks.
\newblock In {\em ICCV}, 2017.

\bibitem{de2018panoptic}
D.~de~Geus, P.~Meletis, and G.~Dubbelman.
\newblock Panoptic segmentation with a joint semantic and instance segmentation
  network.
\newblock {\em arXiv:1809.02110}, 2018.

\bibitem{dvornik2017blitznet}
N.~Dvornik, K.~Shmelkov, J.~Mairal, and C.~Schmid.
\newblock {BlitzNet}: A real-time deep network for scene understanding.
\newblock In {\em ICCV}, 2017.

\bibitem{everingham2015pascal}
M.~Everingham, S.~A. Eslami, L.~Van~Gool, C.~K. Williams, J.~Winn, and
  A.~Zisserman.
\newblock The {PASCAL} visual object classes challenge: A retrospective.
\newblock {\em IJCV}, 2015.

\bibitem{grmi_stuff}
A.~Fathi and K.~Murphy.
\newblock {COCO}-{S}tuff 2017 {C}hallenge: {G-RMI} team.
\newblock 2017.

\bibitem{ghiasi2016laplacian}
G.~Ghiasi and C.~C. Fowlkes.
\newblock Laplacian pyramid reconstruction and refinement for semantic
  segmentation.
\newblock In {\em ECCV}, 2016.

\bibitem{girshick2015fast}
R.~Girshick.
\newblock {Fast R-CNN}.
\newblock In {\em ICCV}, 2015.

\bibitem{girshick2014rcnn}
R.~Girshick, J.~Donahue, T.~Darrell, and J.~Malik.
\newblock Rich feature hierarchies for accurate object detection and semantic
  segmentation.
\newblock In {\em CVPR}, 2014.

\bibitem{Detectron2018}
R.~Girshick, I.~Radosavovic, G.~Gkioxari, P.~Doll\'{a}r, and K.~He.
\newblock Detectron.
\newblock \url{https://github.com/facebookresearch/detectron}, 2018.

\bibitem{he2017mask}
K.~He, G.~Gkioxari, P.~Doll{\'a}r, and R.~Girshick.
\newblock Mask {R-CNN}.
\newblock In {\em ICCV}, 2017.

\bibitem{he2016deep}
K.~He, X.~Zhang, S.~Ren, and J.~Sun.
\newblock Deep residual learning for image recognition.
\newblock In {\em CVPR}, 2016.

\bibitem{honari2016recombinator}
S.~Honari, J.~Yosinski, P.~Vincent, and C.~Pal.
\newblock Recombinator networks: Learning coarse-to-fine feature aggregation.
\newblock In {\em CVPR}, 2016.

\bibitem{hu2017squeeze}
J.~Hu, L.~Shen, and G.~Sun.
\newblock Squeeze-and-excitation networks.
\newblock In {\em CVPR}, 2018.

\bibitem{ioffe2015batch}
S.~Ioffe and C.~Szegedy.
\newblock Batch normalization: Accelerating deep network training by reducing
  internal covariate shift.
\newblock In {\em ICML}, 2015.

\bibitem{kendall2017multi}
A.~Kendall, Y.~Gal, and R.~Cipolla.
\newblock Multi-task learning using uncertainty to weigh losses for scene
  geometry and semantics.
\newblock In {\em CVPR}, 2018.

\bibitem{kirillov2017panoptic}
A.~Kirillov, K.~He, R.~Girshick, C.~Rother, and P.~Doll{\'a}r.
\newblock Panoptic segmentation.
\newblock In {\em CVPR}, 2019.

\bibitem{kirillov2016instancecut}
A.~Kirillov, E.~Levinkov, B.~Andres, B.~Savchynskyy, and C.~Rother.
\newblock {InstanceCut}: from edges to instances with multicut.
\newblock In {\em CVPR}, 2017.

\bibitem{kokkinos2016ubernet}
I.~Kokkinos.
\newblock {UberNet}: Training a universal convolutional neural network for
  low-, mid-, and high-level vision using diverse datasets and limited memory.
\newblock In {\em CVPR}, 2017.

\bibitem{li2018learning}
J.~Li, A.~Raventos, A.~Bhargava, T.~Tagawa, and A.~Gaidon.
\newblock Learning to fuse things and stuff.
\newblock {\em arXiv:1812.01192}, 2018.

\bibitem{li2018weakly}
Q.~Li, A.~Arnab, and P.~H. Torr.
\newblock Weakly-and semi-supervised panoptic segmentation.
\newblock In {\em ECCV}, 2018.

\bibitem{li2016fully}
Y.~Li, H.~Qi, J.~Dai, X.~Ji, and Y.~Wei.
\newblock Fully convolutional instance-aware semantic segmentation.
\newblock In {\em CVPR}, 2017.

\bibitem{lin2016feature}
T.-Y. Lin, P.~Doll{\'a}r, R.~Girshick, K.~He, B.~Hariharan, and S.~Belongie.
\newblock Feature pyramid networks for object detection.
\newblock In {\em CVPR}, 2017.

\bibitem{lin2014coco}
T.-Y. Lin, M.~Maire, S.~Belongie, J.~Hays, P.~Perona, D.~Ramanan,
  P.~Doll{\'a}r, and C.~L. Zitnick.
\newblock Microsoft {COCO}: Common objects in context.
\newblock In {\em ECCV}, 2014.

\bibitem{liu2017sgn}
S.~Liu, J.~Jia, S.~Fidler, and R.~Urtasun.
\newblock {SGN}: Sequential grouping networks for instance segmentation.
\newblock In {\em CVPR}, 2017.

\bibitem{liu2018path}
S.~Liu, L.~Qi, H.~Qin, J.~Shi, and J.~Jia.
\newblock Path aggregation network for instance segmentation.
\newblock In {\em CVPR}, 2018.

\bibitem{liu2016ssd}
W.~Liu, D.~Anguelov, D.~Erhan, C.~Szegedy, S.~Reed, C.-Y. Fu, and A.~C. Berg.
\newblock {SSD}: Single shot multibox detector.
\newblock In {\em ECCV}, 2016.

\bibitem{long2015fully}
J.~Long, E.~Shelhamer, and T.~Darrell.
\newblock Fully convolutional networks for semantic segmentation.
\newblock In {\em CVPR}, 2015.

\bibitem{misra2016cross}
I.~Misra, A.~Shrivastava, A.~Gupta, and M.~Hebert.
\newblock Cross-stitch networks for multi-task learning.
\newblock In {\em CVPR}, 2016.

\bibitem{neuhold2017mapillary}
G.~Neuhold, T.~Ollmann, S.~Rota~Bul\`o, and P.~Kontschieder.
\newblock The mapillary vistas dataset for semantic understanding of street
  scenes.
\newblock In {\em CVPR}, 2017.

\bibitem{newell2016stacked}
A.~Newell, K.~Yang, and J.~Deng.
\newblock Stacked hourglass networks for human pose estimation.
\newblock In {\em ECCV}, 2016.

\bibitem{peng2018megdet}
C.~Peng, T.~Xiao, Z.~Li, Y.~Jiang, X.~Zhang, K.~Jia, G.~Yu, and J.~Sun.
\newblock Megdet: A large mini-batch object detector.
\newblock In {\em CVPR}, 2018.

\bibitem{pham2017biseg}
V.-Q. Pham, S.~Ito, and T.~Kozakaya.
\newblock {BiSeg}: Simultaneous instance segmentation and semantic segmentation
  with fully convolutional networks.
\newblock In {\em BMVC}, 2017.

\bibitem{pinheiro2016learning}
P.~O. Pinheiro, T.-Y. Lin, R.~Collobert, and P.~Doll{\'a}r.
\newblock Learning to refine object segments.
\newblock In {\em ECCV}, 2016.

\bibitem{ren2015a}
S.~Ren, K.~He, R.~Girshick, and J.~Sun.
\newblock {Faster R-CNN}: Towards real-time object detection with region
  proposal networks.
\newblock In {\em NIPS}, 2015.

\bibitem{ronneberger2015u}
O.~Ronneberger, P.~Fischer, and T.~Brox.
\newblock {U-Net}: Convolutional networks for biomedical image segmentation.
\newblock In {\em {MICCAI}}, 2015.

\bibitem{Russakovsky2015}
O.~Russakovsky, J.~Deng, H.~Su, J.~Krause, S.~Satheesh, S.~Ma, Z.~Huang,
  A.~Karpathy, A.~Khosla, M.~Bernstein, A.~C. Berg, and L.~Fei-Fei.
\newblock {ImageNet Large Scale Visual Recognition Challenge}.
\newblock {\em IJCV}, 2015.

\bibitem{tighe2014scene}
J.~Tighe, M.~Niethammer, and S.~Lazebnik.
\newblock Scene parsing with object instances and occlusion ordering.
\newblock In {\em CVPR}, 2014.

\bibitem{tu2005image}
Z.~Tu, X.~Chen, A.~L. Yuille, and S.-C. Zhu.
\newblock Image parsing: Unifying segmentation, detection, and recognition.
\newblock {\em IJCV}, 2005.

\bibitem{wang2018non}
X.~Wang, R.~Girshick, A.~Gupta, and K.~He.
\newblock Non-local neural networks.
\newblock In {\em CVPR}, 2018.

\bibitem{wu2018group}
Y.~Wu and K.~He.
\newblock Group normalization.
\newblock In {\em ECCV}, 2018.

\bibitem{xie2017aggregated}
S.~Xie, R.~Girshick, P.~Doll{\'a}r, Z.~Tu, and K.~He.
\newblock Aggregated residual transformations for deep neural networks.
\newblock In {\em CVPR}, 2017.

\bibitem{yao2012describing}
J.~Yao, S.~Fidler, and R.~Urtasun.
\newblock Describing the scene as a whole: Joint object detection, scene
  classification and semantic segmentation.
\newblock In {\em CVPR}, 2012.

\bibitem{yu2015multi}
F.~Yu and V.~Koltun.
\newblock Multi-scale context aggregation by dilated convolutions.
\newblock In {\em ICLR}, 2016.

\bibitem{zhao2017pspnet}
H.~Zhao, J.~Shi, X.~Qi, X.~Wang, and J.~Jia.
\newblock Pyramid scene parsing network.
\newblock In {\em CVPR}, 2017.

\bibitem{zhao2018psanet}
H.~Zhao, Y.~Zhang, S.~Liu, J.~Shi, C.~C. Loy, D.~Lin, and J.~Jia.
\newblock {PSANet}: Point-wise spatial attention network for scene parsing.
\newblock In {\em ECCV}, 2018.

\bibitem{zhou2017ade20k}
B.~Zhou, H.~Zhao, X.~Puig, S.~Fidler, A.~Barriuso, and A.~Torralba.
\newblock Scene parsing through {ADE20K} dataset.
\newblock In {\em CVPR}, 2017.

\end{thebibliography}

\end{document}